%% file: main.tex
\documentclass[10pt,twocolumn,letterpaper]{article}

\usepackage[pagenumbers]{cvpr} %

\input{preamble}

\definecolor{cvprblue}{rgb}{0.21,0.49,0.74}
\usepackage[pagebackref,breaklinks,colorlinks,allcolors=cvprblue]{hyperref}

\usepackage{graphicx}
\usepackage{caption}
\usepackage{subcaption}
\usepackage{caption} %
\usepackage{adjustbox} %

\def\paperID{-} %
\def\confName{CVPR}
\def\confYear{2026}

\title{\methodname: Offline Feed-Forward 3D Reconstruction at Scale}
\author{Sven Elflein$^{1,2,3}$ \quad Ruilong Li$^{1}$ \quad S{\'e}rgio Agostinho$^{1}$ \\ Zan Gojcic$^{1}$ \quad Laura Leal-Taix{\'e}$^{1}$ \quad Qunjie Zhou$^{1}$ \quad Aljosa Osep$^{1}$ \vspace{0.3em} \\
{\normalsize $^1$NVIDIA} \quad
{\normalsize $^2$Vector Institute} \quad
{\normalsize $^3$University of Toronto}
}

\begin{document}

\twocolumn[{%
\renewcommand\twocolumn[1][]{#1}%
\maketitle
\captionsetup{type=figure}
\begin{center}
\vspace{-0.5cm}

\captionsetup[subfigure]{justification=centering}
\subcaptionbox{Reconstructing of Rome landmarks: Colosseum, Castel Sant'Angelo, Pantheon and Trevi fountain.
\label{fig:scalability_visual}}[.7\textwidth]{%
  
\centering

\includegraphics[width=\linewidth,height=6cm,keepaspectratio]{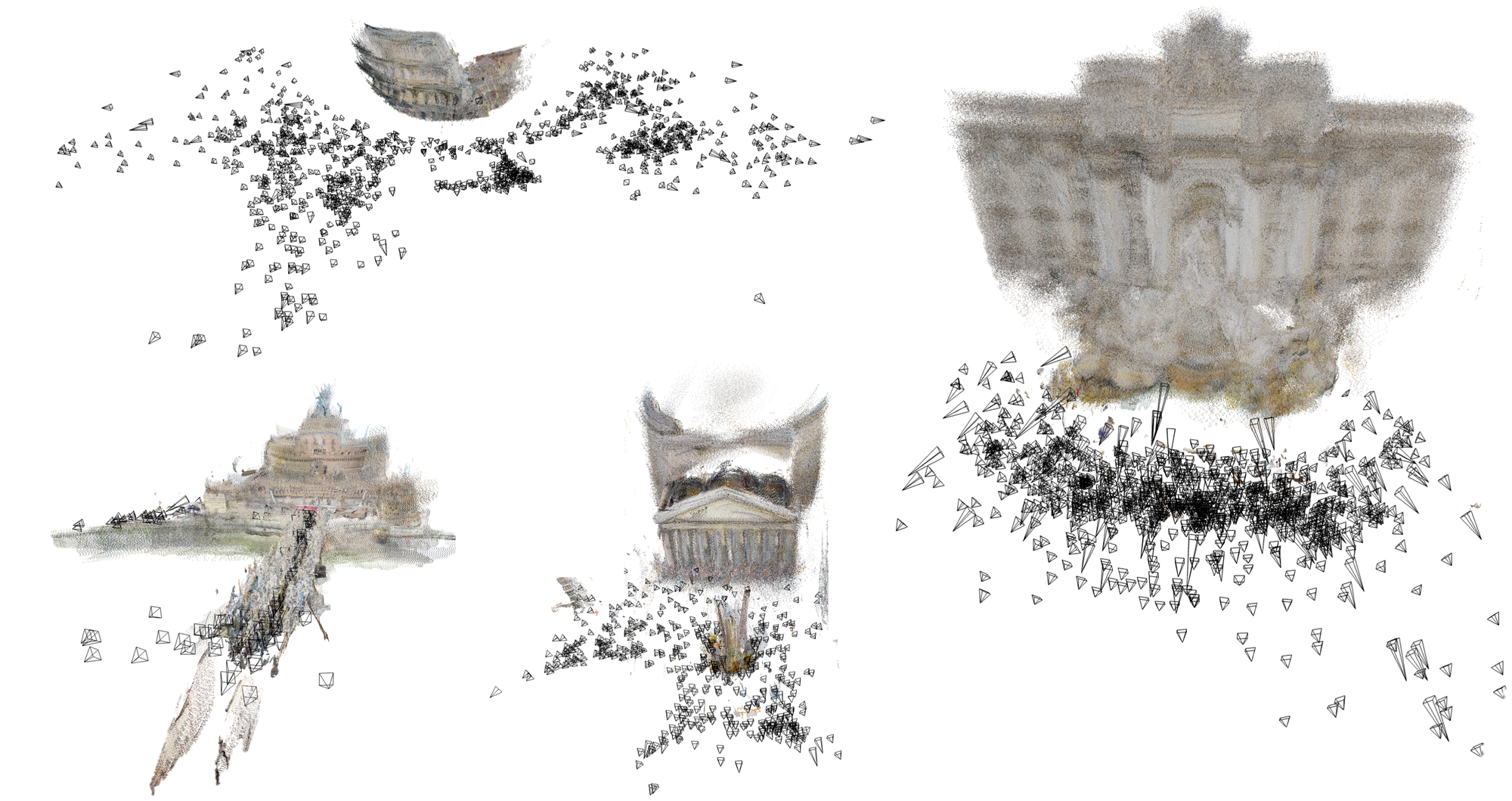}
}\hfill
\subcaptionbox{Num. images \vs inference time. \label{fig:scalability_plot}}[.25\textwidth]{%
  \centering
  \includegraphics[width=\linewidth,height=6cm,keepaspectratio]{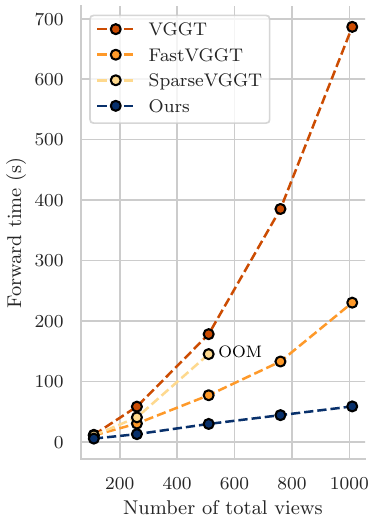}
}
\vspace{-0.2cm}
\caption{
\textbf{Reconstructing Rome landmarks with 1-minute time budget.} We present \methodname, an offline feed-forward 3D reconstruction method that scales linearly \wrt input views (\cref{fig:scalability_plot}). 
As a result, we can reconstruct large scenes from a large number of unposed input views, such as landmarks from tourist-sourced images, in less than a minute via single forward pass (\cref{fig:scalability_visual}).
}
\label{fig:teaser}
\end{center}
}
]
\input{sec/0_abstract}    
\input{sec/1_intro}
\input{sec/2_related_work}

\input{sec/3_method}

\input{sec/4_experiments}

\input{sec/5_conclusion}

\input{sec/6_suppl}
{
    \small
    \bibliographystyle{ieeenat_fullname}
    \bibliography{main}
}

\end{document}

%% file: preamble.tex
\newcommand{\aljnew}[1]{{#1}}

\newcommand{\methodname}{\texttt{VGG-T$^3$}\@\xspace}

\usepackage{xspace}
\usepackage{multirow}
\usepackage{adjustbox}
\usepackage{nicefrac}

\DeclareMathOperator*{\argmin}{arg\,min}

\newcommand{\PAR}[1]{\vskip3pt \noindent {\bf #1~}}
\newcommand{\PARit}[1]{\vskip3pt \noindent {\it #1~}}

\newcommand{\tablestyle}[2]{\setlength{\tabcolsep}{#1}\renewcommand{\arraystretch}{#2}\centering\footnotesize}

%% file: sec/0_abstract.tex
\begin{abstract}
We present a scalable 3D reconstruction model that addresses a critical limitation in offline feed-forward methods: their computational and memory requirements grow quadratically $\wrt$ the number of input images. Our approach is built on the key insight that this bottleneck stems from the varying-length Key-Value (KV) space representation of scene geometry, which we distill into a fixed-size Multi-Layer Perceptron (MLP) via test-time training.
Our \methodname (\textbf{V}isual \textbf{G}eometry \textbf{G}rounded \textbf{T}est \textbf{T}ime \textbf{T}raining) scales linearly $\wrt$ the number of input views, similar to online models, and reconstructs a $1k$ image collection in just $54$ seconds, achieving a $11.6\times$ speed-up over baselines that rely on softmax attention.
Since our method retains global scene aggregation capability, our point map reconstruction error outperforming other linear-time methods by large margins. Finally, we demonstrate visual localization capabilities of our model by querying the scene representation with unseen images.
\end{abstract}

%% file: sec/1_intro.tex
\section{Introduction}
\label{sec:intro}

We tackle large-scale 3D geometry reconstruction from in-the-wild image collections (\cref{fig:scalability_visual}).

\PAR{Status quo.}
Contemporary learning-based approaches directly predict scene geometry from images via feed-forward networks~\cite{wang_dust3r_2024,duisterhof_mast3r-sfm_2025,zhang_flare_2025,wang_vggt_2025,wang_3_2025,keetha_mapanything_2025}.  
These are on-par with classical methods~\cite{snavely_photo_2006,schonberger_structure--motion_2016,pan_global_2024} in terms of accuracy, and are empirically more robust under challenging conditions such as rapid camera motion and low visual overlap. 
However, their computational and memory requirements scale poorly with the number of input images.

This bottleneck originates from the implicit scene-level memory stored in the Key-Value (KV) space of the global self-attention layer. This KV space, projected from all input image tokens, functions as the dense, variable-length scene representation queried for 3D attribute prediction.
To estimate scene geometry from this latent representation, these models need to query the KV space via global softmax attention operations. 
As this operation scales quadratically \wrt the number of input images, 
recent techniques address this issue via sparse attention \cite{wang_faster_2025} or token merging~\cite{shen_fastvggt_2025} to compress the \textit{variable} representation length. 
However, this does not change the underlying quadratic scaling \wrt the number of input images (\cref{fig:scalability_plot}).

\PAR{Compress your KV.} 
A variable-length representation is in contrast to methods that represent the scene geometry via fixed-size implicit representations~\cite{park2019deepsdf, mildenhall_nerf_2020}.
For example, DeepSDF~\cite{park2019deepsdf} conditions a pre-trained decoder on a compact, test-time-optimizable latent code to reconstruct a specific shape conditioned on the observed input 3D point cloud. Intuitively, the fixed-state decoder learns rich geometric priors, while a small latent code encodes instance-specific details through test-time optimization. 
In this work, we revisit this core principle in the context of feed-forward multi-view 3D reconstruction. 

In particular, we leverage a pre-trained multi-view feed-forward model~\cite{wang_vggt_2025} that tokenizes multi-view images and decodes dense depth maps from output tokens.
However, rather than performing (quadratic) softmax attention in the global attention layer, we follow \citet{sun_learning_2025} and map the KV space via weights of a fixed-size MLP. Analogous to DeepSDF, we optimize the MLP at test time with reconstruction loss in \textit{token space}, which allows us to retain the pre-trained encoder/decoder network. 
Querying the KV space at test time to decode depth maps from input views now means only applying the learned MLP to input tokens. This operation is linear \wrt the input image collection size.

\PAR{Large-scale feed-forward reconstruction.} 
With our approach we can perform mini-batching to compute the overall gradient of our test-time training objective.
This implies we can (i) process large image collections on a single GPU by off-loading mini-batches to CPU, and (ii) perform distributed inference by sharding image tokens across multiple GPUs. As a result, we can process a $2k$ image collection in $48.5$s, a $33\times$ improvement over VGGT ($27$min).

\PAR{Visual localization.} Moreover, this representation change also unlocks new capabilities. After reconstructing a set of images, the optimized MLP stores a compressed version of the scene. 
By querying the frozen MLP with a novel query view, we localize this image with respect to the reconstructed scene, thus naturally performing feed-forward visual localization. %
Traditionally, separate solutions are required for reconstruction and localization tasks. In contrast, our approach uses the same model for mapping (optimizing the MLP) and localization (querying the frozen MLP), providing a unified, end-to-end solution. %

\aljnew{
\PAR{To summarize,} \textbf{(i)} we propose an offline feed-forward 3D reconstruction model that scales linearly \wrt the number of input views.
We \textbf{(ii)} show that models that represent scene geometry with variable-length implicit representation (KV) can be ``converted'' into linear-time models via fixed-dimensional implicit state representation.
We \textbf{(iii)} demonstrate our approach supports single-GPU inference for large image sets as well as efficient distributed inference. 
Finally, we \textbf{(iv)} present a proof-of-concept joint feed-forward visual localization and mapping within a single model. 
}

%% file: sec/2_related_work.tex
\section{Related Work}
\label{sec:related_work}

\PAR{Classical pipelines.} Established structure-from-motion techniques such as Bundler~\cite{snavely_photo_2006,snavely_modeling_2008}, COLMAP~\cite{schonberger_structure--motion_2016}, and GLOMAP~\cite{pan_global_2024} follow a multi-stage pipeline that includes feature extraction, correspondence search, camera pose estimation, and joint refinement of camera poses and 3D structure. 
These methods achieve accurate scene reconstructions on large image collections~\cite{agarwal2011building}, provided the scenes are well-constrained (\ie, sufficient visual overlap and connectivity).

\PAR{Feed-forward models.}
Recent feed-forward approaches~\cite{wang_dust3r_2024,leroy_grounding_2024,jang_pow3r_2025} utilize Transformers~\cite{vaswani_attention_2017} to encode input image pairs and regress relative pose and depth maps in the reference view. By encoding spatial relationships via attention mechanisms across image features, these methods can recover 3D geometry, camera motion, and even handle scenes with uncalibrated cameras and low visual overlap. 

\PARit{Multi-view feed-forward methods} encode and aggregate features across multiple views to predict poses and scene geometry simultaneously. Overall, these methods consist of a feature encoder (tokenizer), a multi-view feature aggregator, and a decoder that estimates camera poses and per-view depth maps or global point maps. VGGT~\cite{wang_vggt_2025}, Fast3R~\cite{yang_fast3r_2025}, and $\pi^3$~\cite{wang_3_2025} perform this global fusion in token space via softmax attention. %
Alternatively, Light3R-SfM~\cite{elflein_light3r-sfm_2025} constructs a scene graph from the underlying image collection and pools image features using a shortest-path tree data structure for more efficient aggregation. FLARE~\cite{zhang_flare_2025} decomposes the problem into global camera pose and per-view geometry estimation. %

\PARit{Large-scale reconstruction.} The aforementioned offline multi-view models achieve high accuracy via global self-attention mechanisms, at the cost of quadratic complexity $O(n^2)$ \wrt the number of input views $n$. %
To enable reconstruction with long sequences, Slam3R \cite{liu_slam3r_2025}, VGGT-SLAM \cite{maggio_vggt-slam_2025}, and VGGT-Long \cite{deng_vggt-long_2025} process video data in chunks, using local attention or sliding windows. However, this decouples the global scene state, making them prone to drift and unsuitable for unordered image sets. 
Other methods optimize the global attention operation directly. FastVGGT \cite{shen_fastvggt_2025} uses token merging, and SparseVGGT \cite{wang_faster_2025} employs block-sparse attention. While reducing the constant factor $O(n^2) \to O(\nicefrac{n}{r}^2)$ where $r$ is the token down-sampling ratio, the asymptotic complexity of both approaches remains quadratic. 
These can be viewed as structured compression of the KV scene representation to speed up the global attention operation, with the heuristic that tokens close in image space share similar scene features. 
From this perspective, our work applies flexible compression of the KV space, decoupling the model's computational complexity from the number of input images $n$, thus moving from a quadratic to a linear-time formulation.

\PARit{Online methods.} Several methods process image sequences in an auto-regressive fashion. 
StreamVGGT~\cite{zhuo_streaming_2025} and Stream3R~\cite{lan_stream3r_2025} convert pre-trained VGGT models to causal models, registering newly-observed images into the existing reconstruction, by only attending to prior tokens' keys and values. These methods scale quadratically \wrt $n$ and require memory-intensive KV caching to accelerate inference. %
Online, linear-time models retain past frames/tokens as working memory~\cite{wang_3d_2025}, rely on fixed-size implicit memory that is updated iteratively (CUT3R~\cite{wang_continuous_2025}, Must3R~\cite{cabon_must3r_2025}) or use explicit spatial memory (Point3R~\cite{wu_point3r_2025}, Long3R~\cite{chen_long3r_2025}, MapAnything~\cite{keetha_mapanything_2025}).

\PARit{Test-time training for 3D reconstruction.} 
Concurrent to our work, TTT3R \cite{chen_ttt3r_2025} is an auto-regressive model that builds on the fixed-size memory of CUT3R and reinterprets its internal state update mechanism as test-time training (TTT)~\cite{sun_test-time_2020}. 
Our work utilizes a similar test-time optimization mechanism, but with a fundamentally different interpretation: we resort to test-time optimization to ``compress'' the KV space into a fixed-size MLP. Therefore, our method is global (offline) and, as we show empirically, significantly more accurate compared to TTT3R, yet maintains linear complexity \wrt input size $n$. %

\PAR{Attention with linear complexity.}
The quadratic cost of softmax attention limits scalability in long-sequence modeling. Linear attention methods address this by replacing the softmax kernel with linear feature maps, yielding linear-time, constant-memory recurrences~\cite{kacham2023polysketchformer, katharopoulos_transformers_2020}, with gated~\cite{yang2023gated} and chunkwise-parallel~\cite{hua2022transformer, sun2023retentive, yang_parallelizing_2024, yang2024fla, dao_transformers_2024} extensions improving efficiency and hardware throughput.
State-space models (SSMs) offer an alternative recurrent formulation, where modern variants such as S4~\cite{gu2021efficiently}, H3~\cite{fu2022hungry}, Hyena~\cite{poli_hyena_2023}, and Mamba~\cite{gu2024mamba, dao_transformers_2024} learn structured transitions to capture global dependencies, and can be viewed as gated or structured extensions of linear attention~\cite{behrouz2024titans, yang_gated_2025, sun_learning_2025}.
Recent work shows that TTT provides a strictly more general framework: it treats the hidden state as an optimization variable updated online~\cite{behrouz2024titans, sun_learning_2025, dalal2504one}, recovering linear attention and SSMs as special cases while improving adaptability across domains such as video modeling~\cite{dalal2504one}, novel view synthesis~\cite{zhang_test-time_2025}, and continual learning~\cite{sun_learning_2025}.
A complementary line of work in LLMs explores post-training linearization, converting pretrained transformers into linear-complexity models via lightweight adaptation or distillation~\cite{kasai_finetuning_2021, wang_mamba_2024, mercat_linearizing_2024, dao_transformers_2024, zhang_hedgehog_2024, zhang_lolcats_2025}.
Building on these advances, we extend post-training linearization to multi-view reconstruction, introducing a TTT-based approach that scales bi-directional models to an unbounded number of views.

\PAR{Visual localization.}
The task of localizing a novel query image relative to a pre-built scene representation is typically achieved via geometric correspondence search~\cite{arandjelovic2016netvlad, berton2022cosplace, hausler2021patchvlad, sarlin2019hloc, zhou2022gomatch, panek2022meshloc, sattler2016activesearch}, followed by a Perspective-n-Point (PnP) solver~\cite{ke2017p3p, kneip2011p3p} to compute the final camera pose. 
Similar to ours, Scene Coordinate Regression (SCR) \cite{7scenes, brachmann2021dsac*, brachmann_accelerated_2023, brachmann2024acezero} methods learn a scene-specific function that directly maps input RGB pixels to 3D world coordinates, thereby bypassing the need for explicit feature matching or database queries. 
Recent ACEZero~\cite{brachmann2024acezero} maps and localizes input views jointly from unposed images, however, this end-to-end approach critically relies on extensive, iterative optimization steps to converge toward a stable 3D reconstruction. Our approach instead leverages a pre-trained, feed-forward 3D reconstruction model, which directly enables mapping and localization at test-time with only a few iterations of optimization in token space.

%% file: sec/3_method.tex
\section{Feed-Forward 3D Reconstruction at Scale}
\label{sec:method}

\begin{figure*}[t]
    \centering
    \begin{subfigure}[t]{0.27\textwidth}
        \centering
        \includegraphics[width=\textwidth]{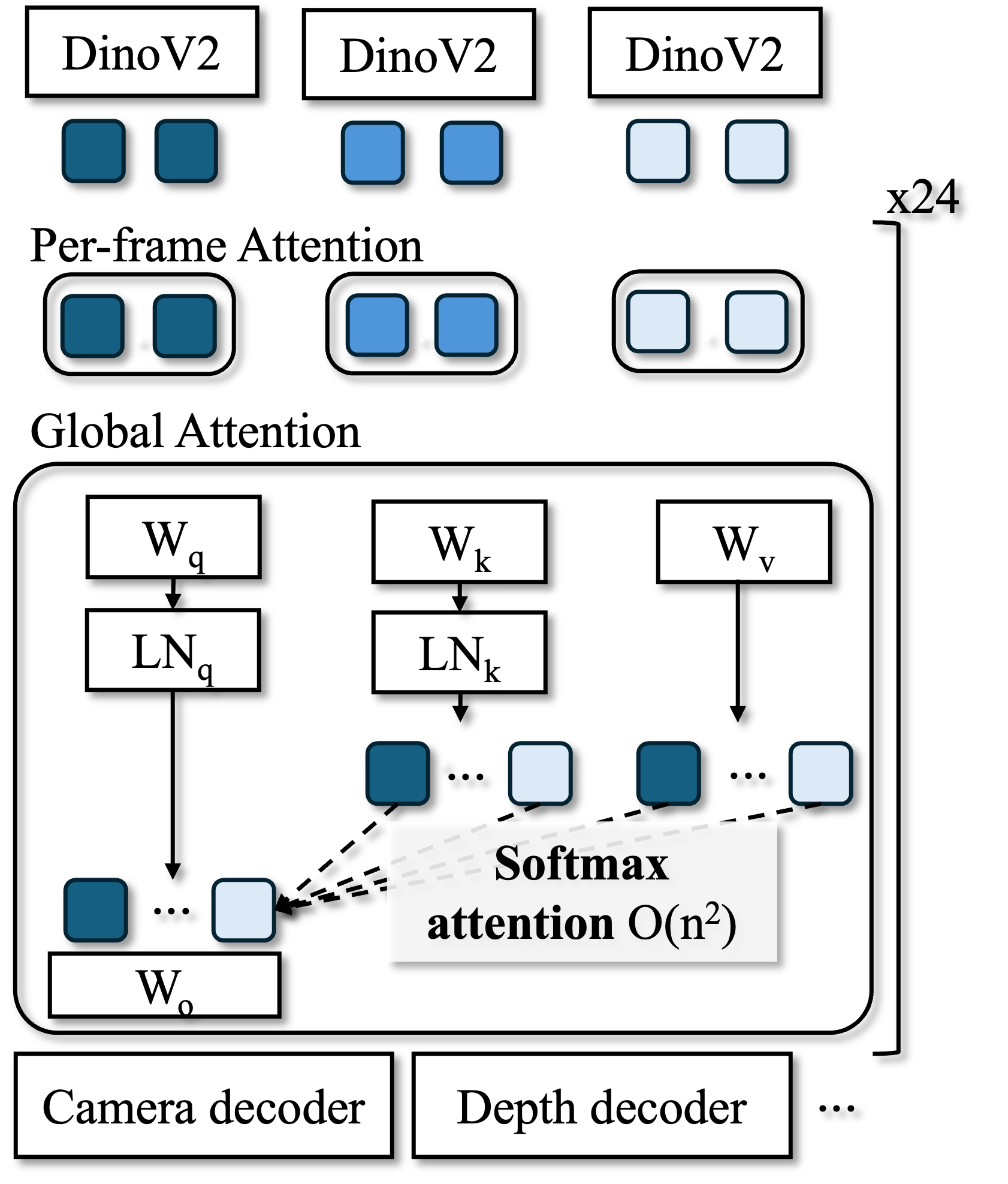}
        \caption{VGGT}
        \label{sec:method:fig:vggt}
    \end{subfigure}
    \hspace{1cm}
    \begin{subfigure}[t]{0.45\textwidth}
        \centering
        \includegraphics[width=\textwidth]{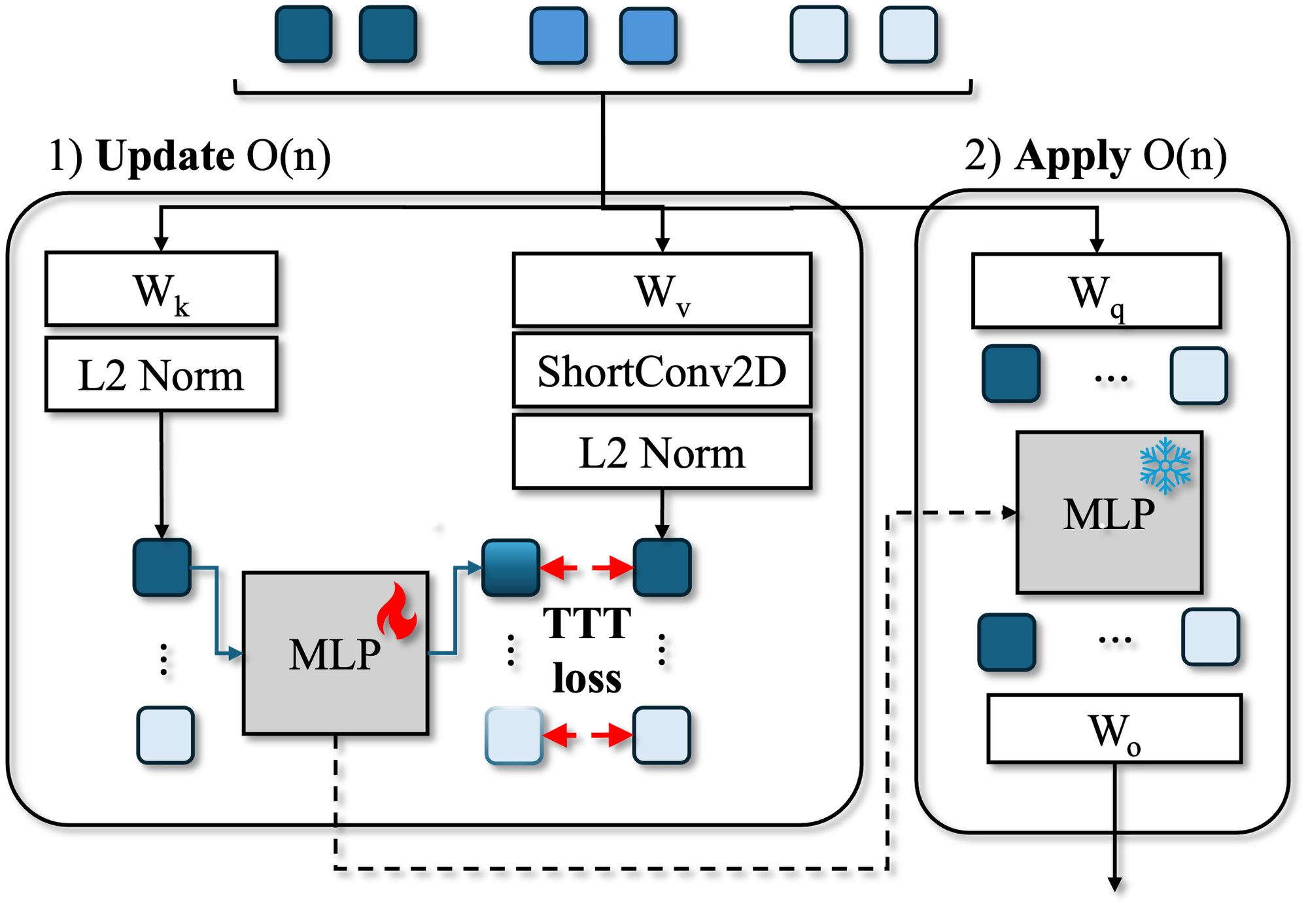}
        \caption{TTT-based global attention replacement with linear scaling.}
        \label{sec:method:fig:ours}
    \end{subfigure}
    \vspace*{-0.1cm}
    \caption{\methodname replaces the global attention block in VGGT (left) with a linear-time alternative based on test-time training (right) to compress the KV space into a fixed-size MLP. We use 3 images for visualization purposes but this scales to arbitrary number of images.}\label{fig:model}
\end{figure*}

We begin by reviewing the recent multi-view feed-forward architecture VGGT~\cite{wang_vggt_2025} and the test-time training (TTT) techniques in~\cref{sec:method:subsec:preliminary}. A key observation is that VGGT implicitly relies on the variable-length key–value (KV) pairs produced by attention layers as its internal representation of scene geometry. While effective, this design requires $O(n^2)$ compute and linearly growing memory with respect to the number of input views.
In \cref{sec:method:subsec:scene_rep}, we introduce our approach, \methodname (\textbf{V}isual \textbf{G}eometry \textbf{G}rounded \textbf{T}est \textbf{T}ime \textbf{T}raining), which replaces these variable-length KV pairs with a compressed, fixed-size MLP representation via TTT. This substitution reduces the computational complexity to $O(n)$, enabling feed-forward 3D reconstruction at scale.

\PAR{Task.}
Given an unordered image collection, denoted as $\{\mathcal{I}_i\}_{i=1}^{N}$ with $\mathcal{I}_i \in \mathbb{R}^{H \times W \times 3}$, the goal is to obtain per-image camera extrinsics $P_i$ consisting of rotation $R \in SE(3)$ and translation vector $T \in \mathbb{R}^3$, pinhole intrinsics $K_i \in \mathbb{R}^{3\times3}$ and dense depth map $X \in \mathbb{R}^{H \times W}$, which represents the geometry observed by individual images.

\subsection{Preliminaries}
\label{sec:method:subsec:preliminary}

\PAR{VGGT.}
VGGT performs multi-view reasoning by first applying an image tokenizer that converts each input image into a sequence of tokens 
$x_i$. It then processes these tokens with attention blocks that alternate between image-wise self-attention and global self-attention across all images (\cref{sec:method:fig:vggt}).
In each attention block, the model projects every input token $x_i$ into query, key, and value (QKV) vectors:
\begin{equation}
\label{sec:method:subsec:prelimary:eq:vggt_proj}
q_i = \text{LN}_q(W_q x_i), k_i = \text{LN}_k(W_k x_i), v_i = W_v x_i. 
\end{equation}
where $W_q$, $W_k$ and $W_v$ are learned linear projections and $\text{LN}_q$ and $\text{LN}_k$ denote layer norms~\cite{ba_layer_2016} performing QK normalization~\cite{dehghani_scaling_2023,henry_query-key_2020} to stabilize training. 
The softmax attention is then applied to obtain per-head output $o_i$ via: 
\begin{equation}
    \label{sec:method:subsec:prelim:seq:sdpa}
    o_i = \sum_j \text{softmax}_j\left(\frac{q_i^T k_j}{\sqrt{d}}\right) v_j.
\end{equation}
Finally, prediction heads operate on the output tokens $o_i$ to directly predict per-image depth, camera poses, and camera intrinsics.
Importantly, the global self-attention layers pool information across all input views, which is essential for multi-view understanding but introduces quadratic complexity w.r.t. the number of views.

\PAR{Test-time training.} 
Recently, \citet{sun_learning_2025} propose to use test-time training~\cite{sun_test-time_2020} in which just a small set of weights $\theta$ (referred to as \textit{Fast weights}~\cite{hinton_using_1987}) are updated \textit{at test-time} using a self-supervised objective $L_{\text{t}}$.
Given queries $q_i$, keys $k_i$, and values $v_i$, \citet{sun_learning_2025} re-define the attention operation as:
\begin{align}
    \label{sec:method:subsec:scene_rep:eq:ttt_opt_1}
    \argmin_{\mathbf{\theta}} & \sum_i L_{t}\bigl( \text{T}_{\mathbf{\theta}}(k_i) - v_i \bigr), \\
    \label{sec:method:subsec:scene_rep:eq:ttt_opt_2}
    o_i &= \text{T}_{\mathbf{\theta}}(q_i).
\end{align}
Intuitively, this optimization embeds the mapping from keys $k_i$ to values $v_i$ into a learnable network $\text{T}_{\mathbf{\theta}}$. Once done, this network can retrieve the appropriate value for a given query $q_i$, analogous to how softmax attention uses QK cosine similarity to retrieve information stored in V (\cref{sec:method:subsec:prelim:seq:sdpa}). Unlike softmax attention, however, both operations in \cref{sec:method:subsec:scene_rep:eq:ttt_opt_1} and \cref{sec:method:subsec:scene_rep:eq:ttt_opt_2} are linear with respect to the sequence length.

\subsection{Can We Fit Rome into MLPs?}
\label{sec:method:subsec:scene_rep}

The central challenge in modern multi-view 3D reconstruction is achieving scalability, which is fundamentally tied to scene representation and its corresponding complexity as the number of input images, $n$, grows. 
Quadratic complexity of VGGT is a direct consequence of variable-length KV scene representation, as extracting information from the KV space scales quadratically \wrt $n$ in the softmax attention~\cite{vaswani_attention_2017} operation, necessary to obtain the output token representation.
\aljnew{This brings us to the core question: \textit{can we bypass the softmax attention operation in KV space?}}

\PAR{Overview.} 
\aljnew{Our method structurally replaces the quadratic global attention operation within the bi-directional Transformer architecture with a linear alternative.
} 
Once we projected the multi-view input to tokens using Transformer-based multi-view networks~\cite{wang_vggt_2025} the forward pass consists of two recurring stages, executed in each global attention layer (\cref{sec:method:fig:ours}):
\textbf{(1) Update}: We project input tokens to queries, keys and values and use TTT~\cite{sun_learning_2025} to compress the variable-length information stored in KV into the fixed-size, compact weights of an MLP. We treat the MLPs as fast weights, \ie, weights that are optimized at train and test time. 
This effectively compresses the key-value mapping of the current layer into a \textit{fixed-size neural} scene representation. 
\textbf{(2) Apply}: 
After optimizing $\theta$, we can query the scene representation efficiently by applying the MLP to queries $q$.
We only apply the MLP in global attention blocks of current layer's queries to obtain updated tokens before they are passed to the next layer. 
Decoding to downstream tasks (per-view depth, ego-pose $P_i$ and camera intrinsics) only occurs after the final Transformer layer.

\PAR{Linearizing the pre-trained model.}
We aim to initialize our model using pre-trained weights of VGGT~\cite{wang_vggt_2025}, including projection matrices $W_q$, $W_k$ and $W_v$ as these already capture general vision knowledge learned by the original model. 
Such linearization is commonly employed in the context of large-language models~\cite{mercat_linearizing_2024} and significantly reduces training cost. 
However, we find empirically that naively applying test-time linearization to replace Softmax Attention (\cref{sec:method:subsec:prelim:seq:sdpa}) with linear-time operation (\cref{sec:method:subsec:scene_rep:eq:ttt_opt_1}) yields very slow convergence during test-time training. 
    
As can be seen in \cref{sec:method:subsec:prelimary:eq:vggt_proj}, token projection involves LayerNorm (LN), which stabilizes Softmax Attention operation in the original model~\cite{henry_query-key_2020}. 
However, LayerNorm involves additional learnable parameters that distort the input space that the MLP is trying to learn at test time. 
By removing LN and instead applying $L_2$ normalization we unlock fast convergence from pre-trained weights.
Moreover, we show that regular softmax attention training followed by our post-training with linearization approach is preferable over directly training from scratch using test-time training.

\PAR{Non-linear spatial mixing.}
While linear attention variants significantly speed up Transformer models, this is generally accompanied by a drop in downstream metrics compared to softmax attention~\cite{gu2024mamba,yang_gated_2025,sun_learning_2025}. 
We attribute this drop in our framework to the inherent mathematical constraints of the TTT objective in \cref{sec:method:subsec:scene_rep:eq:ttt_opt_1}.
Recall that we are learning a mapping from Key to Value space $K \to V$. However, both $K$ and $V$ are derived from same token $x$ via linear projections $K = W_k x$ and $V = W_v x$, and the relationship between them is linear ($V = W_v W_k^{-1} K$, assuming $W_k$ is invertible). Therefore, simply optimizing \cref{sec:method:subsec:scene_rep:eq:ttt_opt_1} can yield a trivial solution. 
To break this dependency and enhance expressivity, we are inspired by the success of sequence mixing layers such as short convolutions~\cite{poli_hyena_2023}, effectively utilized in linear language models~\cite{yang_gated_2025,yang_parallelizing_2024}. 
We adapt this principle for 3D reconstruction by applying \textit{spatial mixing} in the Value ($\mathbf{V}$) space, which we term \textit{ShortConv2D}, forcing the TTT objective to learn a mapping from $K \rightarrow V'$. We implement this as follows:
\begin{enumerate}
    \item \textbf{Reshape}: Given the values $[v_i, \dots, v_{N\times H/p \times W/p}], v_i \in \mathbb{R}^d$, we first reshape the 1D token sequence $\mathbf{V}$ back to its corresponding $2D$ image grid of shape $(N, H/p, W/p, d)$, where $p$ is the tokenizer patch size.
    \item \textbf{Convolve}: We apply a single-layer $2D$ convolution, \textit{ShortConv2D}, which is more suitable for image structures than the $1D$ convolutions typically used in language modeling. This aggregates local neighborhood information to create the context-aware target $\mathbf{V}'$.
    \item \textbf{Flatten}: $\mathbf{V}'$ is reshaped back to a $1D$ sequence before optimizing the TTT objective \cref{sec:method:subsec:scene_rep:eq:ttt_opt_1}.
\end{enumerate}
Intuitively, by applying 2D convolution, the Value $\mathbf{V}'$ for a token now contains aggregated local spatial context, while the Key $\mathbf{K}$ remains context-limited. 
This incentivizes the fast weights optimization (\cref{sec:method:subsec:scene_rep:eq:ttt_opt_1}) to distill a robust geometric scene representation via a stronger self-supervised objective as the MLP must now predict a neighborhood $\mathbf{V}'$ from a single token's feature $\mathbf{K}$.
Concurrently, ViT$^3$~\cite{hanViTUnlockingTestTime2025a} successfully employs convolutions directly in the inner model for the classification task.

\PAR{Test-time scaling.} 
While feed-forward models train on relatively small image collections (up to $24$ in VGGT~\cite{wang_vggt_2025}), our goal is to process significantly larger image collections containing thousands of images, a common requirement in large-scale structure-from-motion~\cite{wilson_robust_2014}. 
While sequence length generalization was studied in the context of softmax attention \cite{jin_training-free_2023}, in the test-time training setting, we observe a large degradation when processing out-of-distribution sequence lengths.
For example, the reconstruction error increases about $5\times$ when extending from $N=100$ to $N=1k$ images of the same scene. 
We hypothesize that the fixed number of optimization steps used during the training (typically one) is insufficient to compress significantly larger scenes to a fixed-dimensional MLP. To confirm, we log the top-performing optimizer step of the TTT objective (\cref{sec:method:subsec:scene_rep:eq:ttt_opt_1}) across two scales: $20$ images (in-distribution) and $1k$ images (out-distribution, a $\sim 50\times$ increase).

\begin{figure}[t]
    \centering
    \begin{subfigure}{0.48\linewidth}
        \centering
        \includegraphics[width=\linewidth]{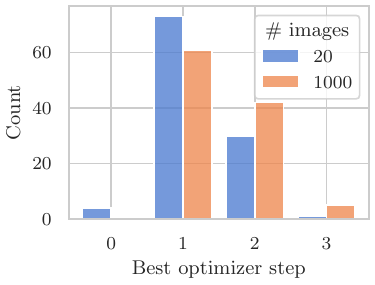}
        \subcaption{Best number of optimizer step of test-time training objective for two sizes of image collections.}
        \label{sec:method:fig:best_optimizer_step_seq_length}
    \end{subfigure}
    \hfill
    \begin{subfigure}{0.48\linewidth}
        \centering
        \includegraphics[width=\linewidth]{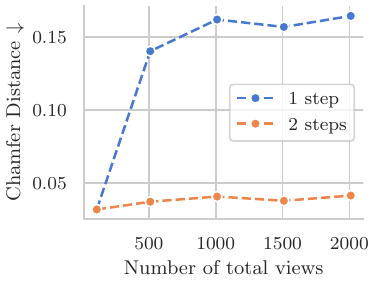}
        \subcaption{Pointmap prediction error for different number of input images 
        (lower is better).}
        \label{sec:method:fig:pointmap_err_seq_length}
    \end{subfigure}
    \caption{Sequence-length generalization analysis.}
    \label{sec:method:fig:sequence_length_scaling}
\end{figure}

As can be seen in \cref{sec:method:fig:best_optimizer_step_seq_length}, for in-distribution samples, one step is sufficient, while for $1k$ images, it is beneficial to increase number of optimizer steps. Simply performing more optimizer steps, we achieve almost constant scaling to arbitrary sequence lengths, showing a form of test-time scaling via additional computation~\cite{deepseek-ai_deepseek-r1_2025}. As increasing the number of steps further does not aid reconstruction quality we perform $2$ steps unless otherwise noted.

\subsection{Large-scale Reconstruction}
\label{sec:method:subsec:implication}
We discuss the implications of our scene representation change in the following.

\PAR{Scalability.} Comparing \cref{sec:method:subsec:prelim:seq:sdpa} with \cref{sec:method:subsec:scene_rep:eq:ttt_opt_1}, we first note that the complexity of the operations changes from $O(n^2)$ to $O(n)$, resolving the quadratic bottleneck within the global attention layers found in existing reconstruction models.

\PAR{Flexible inference strategies.} Moreover, this change unlocks inference strategies that enable us to process arbitrarily large image collections on a single GPU and accelerate throughput linearly on multiple GPUs via distributed inference \aljnew{by applying TTT in a minibatch fashion.}

Recall that TTT optimization learns the MLP weights $\theta$ to map local features $K \to V$, and the loss function (\cref{sec:method:subsec:scene_rep:eq:ttt_opt_1}) is sum over all input tokens $i$. 
As the overall optimization objective is simply sum of local losses, the total gradient of the loss \wrt the MLP weights $\theta$ is also a sum of local gradients (\cref{sec:method:subsec:implication:eq:linearity_aljosa}):
\begin{equation}
\label{sec:method:subsec:implication:eq:linearity_aljosa}
\frac{d L_{\text{total}}}{d\theta} = \sum_i \frac{d}{d\theta} L(\mathbf{k}_i, \mathbf{v}_i) = \sum_s \left( \sum_{i \in s} \frac{d}{d\theta} L(\mathbf{k}_i, \mathbf{v}_i) \right).
\end{equation}
As noted in \citet{zhang_test-time_2025}, this implies we can compute the gradient of the objective on \aljnew{minibatches} $s$ independently. 
For distributed inference, we can process \aljnew{minibatches} on individual GPUs, and synchronize gradients. This enables efficient training in cases where the sequence does not fit into memory of a single GPU.
In practice we shard images such that each GPU only processes a subset $s$. 
In the global layers, we then use \cref{sec:method:subsec:implication:eq:linearity_aljosa} to synchronize the MLP weights across GPUs by performing all-to-all communication which is efficient due to their small size. 

Moreover, this property also allows processing arbitrarily large image collections on a single GPU. For this, we off-load \aljnew{minibatches} (\cref{sec:method:subsec:implication:eq:linearity_aljosa}) to host memory (instead of distributing across GPUs). %
We can then compute the update for the entire sequence by loading a \aljnew{minibatch} at a time to device memory, compute the gradient, and off-load \aljnew{minibatch} back to host memory. This requires keeping only a single \aljnew{minibatch} in device memory at a time. 
Note that methods relying on softmax attention (\eg, VGGT~\cite{wang_vggt_2025} and its sparse variants~\cite{shen_fastvggt_2025,wang_faster_2025} using FlashAttention~\cite{dao_flashattention_2022})
require $q_i, k_i, v_i$ of all images to be in GPU memory which, even for large GPUs, leads quickly to out-of-memory errors when processing larger image collections.

\PAR{Query-able reconstruction \& visual localization.} 
After processing a set of images representing a scene, our network can be queried with new observations. It outputs scene geometry and camera pose of the new image relative to the existing reconstruction. 
To do so, we keep test-time optimized weights frozen, and run standard forward pass for a new query image, with one key modification: in the global attention layers, we only apply the frozen MLPs to the query features $q_i$ to retrieve information from the scene representation, without updating the MLP parameters $\theta$. 
This effectively transforms the model into a single-image Transformer for query processing.
We show in \cref{sec:experiments:subsec:visloc} that this querying mechanism enables us to perform visual localization. %

%% file: sec/4_experiments.tex
\section{Experiments}
\label{sec:experiments}

In this section, we compare our \methodname to state-of-the-art offline and online baselines on standard tasks and benchmarks, examining accuracy \vs runtime in both the conventional setting (\cref{sec:experiments:subsec:standard_benchmarks}) and a large-scale regime (\cref{sec:experiments:subsec:large_scale}). We further demonstrate that our approach enables accurate feed-forward 3D visual localization in unposed, in-the-wild image collections (\cref{sec:experiments:subsec:visloc}). Finally, we present ablation studies that validate our design choices in \cref{sec:experiments:subsec:ablations}.

\PAR{Implementation details.}
We start from the public VGGT checkpoint and convert it to a linearized model by replacing all global attention layers with TTT layers. Following LaCT~\cite{zhang_test-time_2025}, our TTT layer uses a SwiGLU MLP~\cite{shazeer_glu_2020} to learn the $K \rightarrow V$ mapping, Muon~\cite{jordan2024muon} for optimization, and dot product loss $L_t(T_\theta(k_i), v_i) = T_\theta(k_i)^T v_i $. After all the QKV projection layer, we additionally apply an $3\times3$ \emph{ShortConv2D} on $V$ for non-linear spatial mixing.
We freeze all original VGGT parameters and fine-tune only global attention layers using a dataset comparable to VGGT’s original training data, running for $100k$ steps on 8 NVIDIA A100-80GB ($\approx$ 12\% of the cost of training VGGT from scratch). 
For more details we refer to the appendix.

\PAR{Baselines.}
We compare our approach with both offline and online reconstruction methods. On the offline side, we include VGGT~\cite{wang_vggt_2025} as an upper bound in reconstruction accuracy, along with efficient variants such as FastVGGT~\cite{shen_fastvggt_2025} and SparseVGGT~\cite{wang_faster_2025}, all of which exhibit quadratic complexity with respect to the number of input views. On the online side, we benchmark against TTT3R~\cite{chen_ttt3r_2025}, a concurrent method that improves upon CUT3R~\cite{wang_continuous_2025} and is designed for ordered input sequences with linear complexity. We carefully analyze the accuracy–scalability trade-offs of these baselines alongside our method.

\begin{table*}[t]
    \centering
    \scriptsize
    \input{tables/pointmap_standard.tex}
    \caption{\textbf{Pointmap estimation on dense (\textit{-D}) and sparse (\textit{-S}) split.} Overall, we outperform $O(n)$ baseline, TTT3R, and remain competitive \wrt $O(n^2)$ baselines. FastVGGT code fails on NRGBD-S due to one instance having only two views.
    }
    \label{tab:pointmap_standard}
\end{table*}

\begin{table}[t]
    \centering
    \resizebox{\linewidth}{!}{%
    \input{tables/video_depth}
    }
    \caption{\textbf{Video depth estimation.} \methodname outperforms sequential $O(n)$ baseline by a substantial margin and performs on-par with $O(n^2)$ baselines.
    }
    \label{tab:videodepth_standard}
\end{table}

\begin{table}[t]
    \centering
    \tablestyle{1pt}{1.05}
    \resizebox{\linewidth}{!}{%
    \input{tables/camera_poses}
    }
    \caption{\textbf{Camera pose estimation.}
    Our method supports both ordered and unordered input sequences, whereas the other TTT3R performs poorly on unordered inputs. Via sequential processing, TTT3R provides more accurate pose estimates. Best performance on ordered inputs are marked \textbf{bold}, best un-ordered \textcolor{blue}{blue}.
    }
    \label{tab:camera_poses_standard}
\end{table}

\subsection{Standard Benchmarks}
\label{sec:experiments:subsec:standard_benchmarks}

We thoroughly validate our implications by evaluating our method on the three common geometric downstream tasks, \ie, pointmap estimation, video depth and camera pose estimation  with their standard benchmarks.

\PAR{Pointmap estimation.}
Following prior work~\cite{wang_continuous_2025,wang_3d_2025}, we evaluate multi-view point-map estimation on NRGBD~\cite{azinovic_neural_2022}, 7scenes~\cite{7scenes}, DTU~\cite{jensen_large_2014}, and ETH3D~\cite{schops_multi-view_2017}, using Chamfer Distance and Normal Consistency~\cite{wang_3d_2025} to assess the quality of the reconstructed points and surfaces, respectively.
As shown in \cref{tab:pointmap_standard}, we outperform the other $O(n)$ baseline, TTT3R, on all benchmarks except CD on 7scenes-D, where we are only marginally worse. Notably, our method reduces error by $2-2.5\times$ on DTU, ETH3D, and NRGBD-D. Compared to $O(n^2)$ baselines, we remain competitive and even surpass their performance on DTU.

\PAR{Video depth.}
Following \cite{wang_continuous_2025}, we also report the performance on the task of video depth estimation using Bonn~\cite{palazzolo_refusion_2019}, KITTI~\cite{geiger_vision_2013} and Sintel~\cite{butler_naturalistic_2012} evaluation sets. As in prior work,
we align predictions using a single scale per sequence and report the Absolute Relative Error (Abs. Rel.) as well as the percentage of predictions with $\delta < 1.25$. As shown in~\cref{tab:videodepth_standard}, our method outperforms the $O(n)$ baseline TTT3R on two of the three datasets—by a substantial margin and achieves performance on par with $O(n^2)$ methods on KITTI dataset.

\PAR{Camera pose estimation.}
We further evaluate our model on the task of camera pose estimation using TUM-RGBD~\cite{sturm_benchmark_2012}, ScanNet~\cite{dai_scannet_2017}, and Sintel~\cite{butler_naturalistic_2012}. While our method shows consistent advantages on other tasks, we observe that our TTT-linearized model struggles on camera pose estimation, as shown in \cref{tab:camera_poses_standard}. We suspect this is related to VGGT’s special treatment of camera pose, where a dedicated camera token is appended to the image tokens immediately before the attention layer, effectively creating two input ``modalities''. This heterogeneous structure may be challenging for the MLP within the TTT layer to memorize, which highlights an interesting direction for future research.
Nevertheless, it is worth noting that our method naturally supports both ordered and unordered input sequences, whereas the other $O(n)$ baseline, TTT3R, degrades under unordered inputs, as shown in \cref{tab:camera_poses_standard}.

\subsection{Large-Scale 3D Reconstruction}
\label{sec:experiments:subsec:large_scale}

As discussed in \cref{sec:method:subsec:implication}, our method preserves the accuracy advantages of offline, global reconstruction while scaling linearly with the number of input views, thereby enabling large-scale 3D scene reconstruction.

\PAR{Setup.}
To evaluate the scalability of each model, we use the 7scenes dataset, which provides sufficient video coverage for large-scale reconstruction. For each scene, we aggregate all video frames and uniformly subsample the images to form the validation set. All remaining implementation details and evaluation metrics follow \cref{sec:experiments:subsec:standard_benchmarks}.

\begin{figure}[t]
    \centering
    \includegraphics[width=1.0\linewidth]{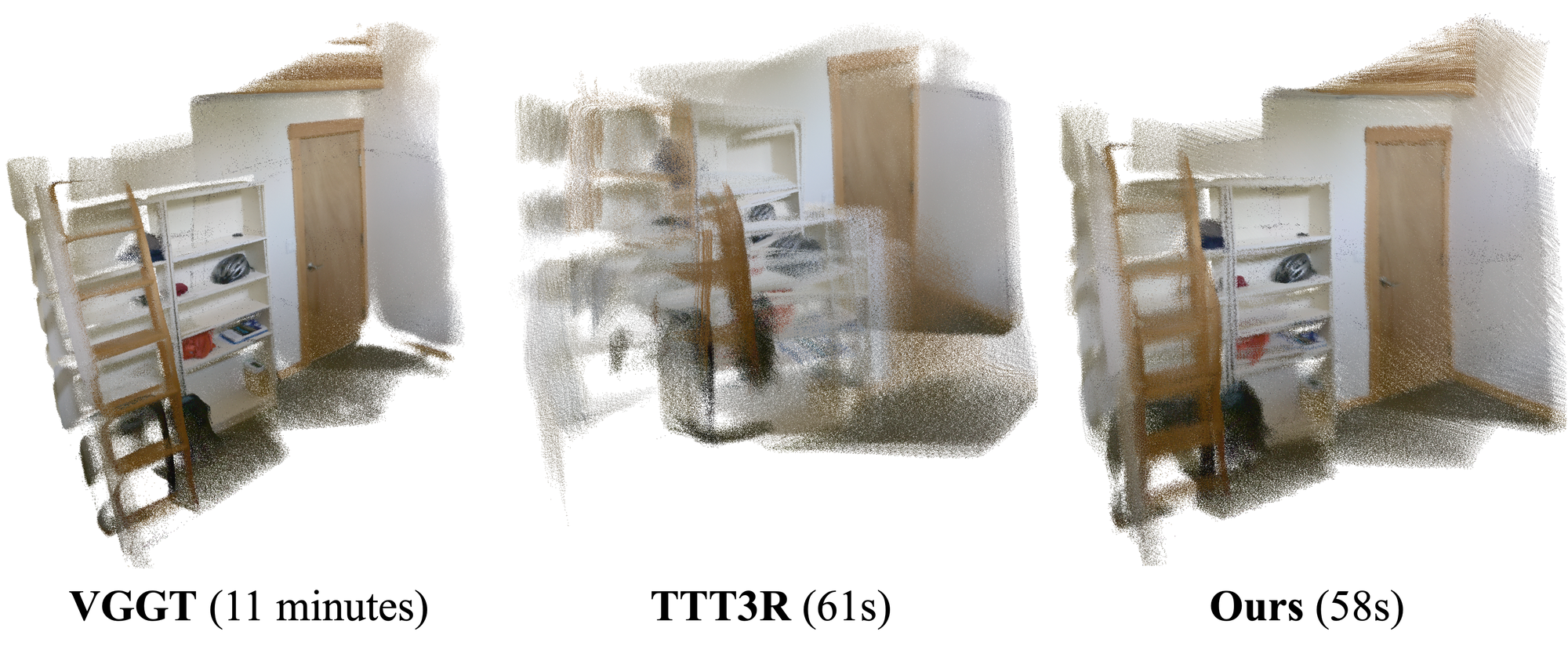}
    \vspace*{-0.5cm}
    \label{sec:experiments:fig:qualitative}
\end{figure}

\begin{figure}[t]
    \centering
    \includegraphics[width=0.8\linewidth]{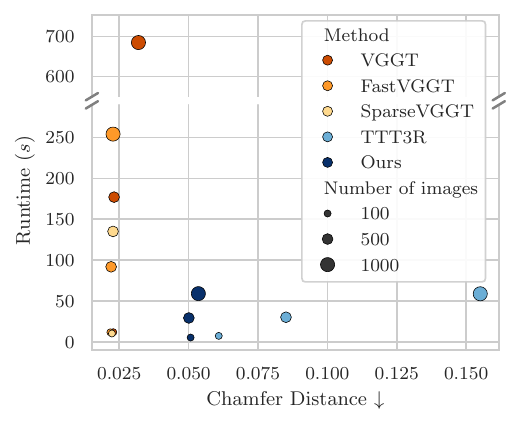}
    \caption{\textbf{Runtime ($\downarrow$) \vs Chamfer distance ($\downarrow$) for collections of size $\in \{ 100, 500, 1k\}$ on 7scenes dataset.} 
    In terms of reconstruction quality (Chamfer distance), we observe a small gap between \methodname and $O(n^2)$ baselines, that narrows with increasing number of images. However, for $1k$ input, VGGT takes ca. $11$min while \methodname only needs $58$ seconds ($11.6\times$ speedup). \methodname scales comparably to TTT3R and does not degrade \wrt increasing number of images. 
    }
    \label{sec:experiments:fig:scaling_metric_runtime}
\end{figure}

\PAR{Results.} 
We report runtime and reconstruction quality with different image collection sizes in \cref{sec:experiments:fig:scaling_metric_runtime}. 
Comparing to $O(n^2)$ methods such as VGGT, FastVGGT, and SparseVGGT, \methodname scales linearly and thus is substantially faster: it reconstructs $1k$ images in 58 seconds, whereas VGGT requires over 11 minutes ($11.6 \times$ slower) and FastVGGT takes more than 4 minutes ($4.3\times$ slower). Comparing to the state-of-the-art $O(n)$ alternative, TTT3R, \methodname delivers significantly higher reconstruction accuracy and maintains stable performance even when scaling to image counts far beyond those seen during training. A visual comparison can be found in \cref{sec:experiments:fig:qualitative}.

\PAR{Distributed inference.}
Our method naturally supports multi-GPU distributed inference for additional speedup, as shown in \cref{tab:dist_inference} and discussed in \cref{sec:method:subsec:implication}. In contrast to VGGT, which requires carefully engineered context-parallel implementations for softmax attention (e.g., ring attention~\cite{liu_ring_2023}), \methodname works directly with distributed data parallel (DDP) as cross-GPU communication is only needed during the fast-weight (MLP) update. The alternative $O(n)$ method, TTT3R, is not compatible with multi-GPU inference due to its autoregressive processing.

\begin{table}[t]
    \centering
    \resizebox{\linewidth}{!}{%
    \input{tables/dist_inference}
    }
    \caption{\textbf{Reconstruction latency (s) with distributed inference.} \methodname can efficiently process large sequences on a single GPU, and provide linear speed-up via distributed inference.}
    \label{tab:dist_inference}
\end{table}

\subsection{Feed-forward Visual Localization}
\label{sec:experiments:subsec:visloc}

As discussed in \cref{sec:method:subsec:implication}, we can query our model with new images that were not part of test-time optimization. This can be interpreted as feed-forward visual localization with respect to the implicit map produced via TTT.

\PAR{Setup.} We evaluate our approach on two commonly used datasets: 7scenes~\cite{7scenes} and Wayspots~\cite{brachmann_accelerated_2023,arnold_map-free_2022}, and compare to TTT3R which, as an autoregressive model, also maintains a state that can be queried in $O(1)$. 
We perform Sim(3) alignment between the ground truth and predicted poses for the mapping images, then measure the rotation $e_r$ and translation error $e_t$ of the predicted query image poses as well as the percentage of query images localized within thresholds $e_r < T_r, e_t < T_t$.

\PAR{Discussion.} As shown in \cref{tab:visloc}, \methodname outperforms TTT3R on both benchmarks, with particularly large improvements on Wayspots. This demonstrates that our MLP-based scene representation enables effective feed-forward visual localization.
We note that the state-of-the-art visual localization pipelines that utilize accurate camera poses during explicit mapping, could achieve more accurate localization, \eg, Reloc3R~\cite{dong_reloc3r_2025} achieves $e_r=1.02^\circ, e_t=0.04\text{m}$ on 7scenes -- our aim is to show that feed-forward visual localization without explicit mapping is indeed feasible, and opens exciting future research directions.

\begin{table}[t]
    \centering
    \adjustbox{max width=\linewidth}{%
    \input{tables/visloc}
    }
    \caption{\textbf{Feed-forward visual localization in unposed image collection.} The MLP-based state representation in \methodname allows for more precise localization of new images compared to TTT3R.}
    \label{tab:visloc}
\end{table}

\subsection{Ablations}
\label{sec:experiments:subsec:ablations}

In \cref{tab:ablation} we outline key ablation studies that justify our design choices, that we perform in a smaller scale setting with image resolution of $224\times224$ on ScanNet++ training with $2-24$ views. 
All models use the same base architecture.

\PAR{Variants.}
We train a model using softmax attention as a upper-bound performance reference. We apply our linearization approach to this model as described in \cref{sec:method:subsec:scene_rep}.
As a baseline for linearization we adapt \textit{T2R}~\cite{kasai_finetuning_2021} to our setting. 
We also consider a variant where linearize the model without initialization from pre-trained weight. 
Finally, we ablate the effectiveness of adding ShortConv2D as discussed in \cref{sec:method}.

\begin{table}[t]
    \centering
    \adjustbox{max width=0.8\linewidth}{%
    \input{tables/ablation.tex}
    }
    \caption{\textbf{Ablations.} We evaluate key design decisions behind our linearization and ShortConv2D design.} 
    \label{tab:ablation}
\end{table}

\PAR{Discussion.}
As reported in \cref{tab:ablation}, we find that training with TTT from scratch (i) gets stuck in a local optimum and linearizing the model pre-trained with softmax atttention is key for good performance. Our linearization (iii) with test-time training significantly outperforms T2R~\cite{kasai_finetuning_2021} (ii) and LoLCats~\cite{zhang_lolcats_2025} (iii). Finally, using ShortConv2D (v) further closes the gap towards softmax attention.

%% file: tables/pointmap_standard.tex
\begin{tabular}{llrrrrrrrrrrrr}
\toprule
 &  & \multicolumn{2}{c}{\textbf{7scenes-D}} & \multicolumn{2}{c}{\textbf{7scenes-S}} & \multicolumn{2}{c}{\textbf{DTU}} & \multicolumn{2}{c}{\textbf{ETH3D}} & \multicolumn{2}{c}{\textbf{NRBGD-D}} & \multicolumn{2}{c}{\textbf{NRBGD-S}} \\
\cmidrule(lr){3-4} \cmidrule(lr){5-6} \cmidrule(lr){7-8} \cmidrule(lr){9-10} \cmidrule(lr){11-12} \cmidrule(lr){13-14} 
 &  & CD $\downarrow$ & NC $\uparrow$ & CD $\downarrow$ & NC $\uparrow$ & CD $\downarrow$ & NC $\uparrow$ & CD $\downarrow$ & NC $\uparrow$ & CD $\downarrow$ & NC $\uparrow$ & CD $\downarrow$ & NC $\uparrow$ \\
\midrule
\multirow[c]{3}{*}{$O(n^2)$} & VGGT & 0.024 & 0.668 & 0.054 & 0.731 & 1.537 & 0.676 & 0.279 & 0.855 & 0.014 & 0.897 & 0.055 & 0.894 \\
 & SparseVGGT & 0.023 & 0.665 & 0.094 & 0.694 & 1.541 & 0.675 & 0.327 & 0.836 & 0.018 & 0.876 & 0.079 & 0.859 \\
 & FastVGGT & 0.021 & 0.662 & 0.065 & 0.719 & 1.683 & 0.672 & 0.594 & 0.775 & 0.033 & 0.841 & - & - \\
\midrule
\multirow[c]{2}{*}{$O(n)$} & TTT3R & 0.035 & 0.666 & 0.129 & 0.685 & 5.708 & 0.672 & 0.885 & 0.733 & 0.071 & 0.811 & 0.094 & 0.819 \\
 & \methodname (ours) &  \textbf{0.030} & \textbf{0.679} & \textbf{0.107} & \textbf{0.716} & \textbf{1.654} & \textbf{0.685} & \textbf{0.480} & \textbf{0.789} & \textbf{0.029} & \textbf{0.867} & \textbf{0.056} & \textbf{0.893} \\
\bottomrule
\end{tabular}

%% file: tables/video_depth.tex
\begin{tabular}{lcccccc}
\toprule
 & \multicolumn{2}{c}{\textbf{Bonn}} & \multicolumn{2}{c}{\textbf{KITTI}} & \multicolumn{2}{c}{\textbf{Sintel}} \\
\cmidrule(lr){2-3} \cmidrule(lr){4-5} \cmidrule(lr){6-7} 
 & $\delta < 1.25$ $\uparrow$ & Abs Rel $\downarrow$ & $\delta < 1.25$ $\uparrow$ & Abs Rel $\downarrow$ & $\delta < 1.25$ $\uparrow$ & Abs Rel $\downarrow$ \\
\midrule
VGGT & 0.967 & 0.059 & 0.964 & 0.071 & 0.646 & 0.300 \\
SparseVGGT & 0.968 & 0.057 & 0.963 & 0.070 & 0.639 & 0.304 \\
FastVGGT & 0.969 & 0.058 & 0.964 & 0.073 & 0.630 & 0.307 \\
\midrule
TTT3R & \textbf{0.969} & \textbf{0.061} & 0.818 & 0.151 & 0.510 & 0.469 \\
\methodname (ours) & 0.963 & 0.063 & \textbf{0.967} & \textbf{0.076} & \textbf{0.581} & \textbf{0.345} \\
\bottomrule
\end{tabular}

%% file: tables/camera_poses.tex
\begin{tabular}{lccccccccc}
\toprule
& \multicolumn{3}{c}{ScanNet} & \multicolumn{3}{c}{Sintel} & \multicolumn{3}{c}{TUM} \\
\cmidrule(lr){2-4} \cmidrule(lr){5-7} \cmidrule(lr){8-10}
& ATE $\downarrow$ & $\text{RPE}_r$ $\downarrow$ & $\text{RPE}_t$ $\downarrow$ & ATE $\downarrow$ & $\text{RPE}_r$ $\downarrow$ & $\text{RPE}_t$ $\downarrow$ & ATE $\downarrow$ & $\text{RPE}_r$ $\downarrow$ & $\text{RPE}_t$ $\downarrow$ \\
\midrule
VGGT & 0.035 & 0.381 & 0.016 & 0.172 & 0.467 & 0.061 & 0.012 & 0.310 & 0.010 \\
SparseVGGT & 0.036 & 0.394 & 0.016 & 0.177 & 0.552 & 0.070 & 0.013 & 0.316 & 0.010 \\
FastVGGT & 0.035 & 0.494 & 0.018 & 0.158 & 0.516 & 0.060 & 0.013 & 0.317 & 0.011 \\
\midrule
TTT3R & \textbf{0.063} & \textbf{0.617} & \textbf{0.020} & \textbf{0.196} & \textbf{0.767} & \textbf{0.088} & \textbf{0.025} & \textbf{0.337} & \textbf{0.012} \\
TTT3R (unordered) & 0.094 & 3.942 & 0.089 & 0.325 & \textcolor{blue}{1.462} & 0.213 & \textcolor{blue}{0.029} & 0.652 & 0.029 \\
\methodname (ours) & \textcolor{blue}{0.070} & \textcolor{blue}{0.878} & \textcolor{blue}{0.034} & \textcolor{blue}{0.234} & 1.553 & \textcolor{blue}{0.117} & 0.037 & \textcolor{blue}{0.533} & \textcolor{blue}{0.028} \\
\bottomrule
\end{tabular}

%% file: tables/dist_inference.tex
\begin{tabular}{lrrr|rrr}
    \toprule
     & \multicolumn{3}{c|}{1500 images} & \multicolumn{3}{c}{2000 images} \\
     & 1 GPU & 2 GPUs & 4 GPUs & 1 GPU & 2 GPUs & 4 GPUs \\
    \midrule
    TTT3R & 90.1 & N/A & N/A & 126.2 & N/A & N/A \\
    VGGT & OOM & 1779.3 & 913.6 & OOM & 2827.1 & 1590.2 \\
    \methodname (ours) & 173.1 & 56.8 & \textbf{29.7} & 230.7 & 74.8 & \textbf{48.5} \\
    \bottomrule
\end{tabular}

%% file: tables/visloc.tex
\begin{tabular}{llrrrr}
\toprule
 &  & $e_r (^{\circ})$ $\downarrow$ & $e_t$ (m) $\downarrow$ & $10$cm, $10^{\circ}$ (\%) $\uparrow$ & $20$cm, $20^{\circ}$ (\%) $\uparrow$ \\
\midrule
\multirow[c]{2}{*}{7Scenes} & TTT3R & 7.18 & 0.17 & 34.59 & 70.21 \\
 & Ours & \textbf{6.71} & \textbf{0.16} & \textbf{40.69} & \textbf{73.00} \\
 \midrule
\multirow[c]{2}{*}{Wayspots} & TTT3R & 74.45 & 4.38 & 0.69 & 2.94 \\
 & Ours & \textbf{32.04} & \textbf{1.90} & \textbf{13.41} & \textbf{30.64} \\
\bottomrule
\end{tabular}

%% file: tables/ablation.tex
\begin{tabular}{lrrr}
\toprule
 & CD $\downarrow$ & NC $\uparrow$ & mAA(30) $\uparrow$ \\
\midrule
Softmax Attention & 0.061 & 0.844 & 76.33 \\
\midrule
(i) Scratch & 0.262 & 0.727 & 52.95 \\
(ii) T2R~\cite{kasai_finetuning_2021} & 0.137 & 0.804 & 66.27 \\
(iii) LoLCats~\cite{zhang_lolcats_2025} & 0.097 & 0.804 & 62.87 \\ 
(iv) Ours & 0.074 & 0.833 & 72.16 \\
(v) Ours + ShortConv2D & \textbf{0.066} & \textbf{0.838} & \textbf{74.14} \\
\bottomrule
\end{tabular}

%% file: sec/5_conclusion.tex
\section{Conclusion}
\label{sec:conclusion}

We presented scalable feed-forward 3D reconstruction that gracefully scales with the number of input views. At the core of our approach is learning a mapping from Keys to Values via test-time optimization instead of querying the KV representation with softmax attention, an operation that scales quadratically \wrt number of input views. Our efficient linearization of the VGGT model allows reconstruction of $1k$ images $11.6\times$ and $2k$ images up to $33\times$ faster while outperforming linear-time methods on pointmap and video depth estimation by large margins. 
\PAR{Limitations.} 
Empirically, we show that our approach retains scalability of online (auto-regressive) methods, and provides a significantly more accurate depth and point maps due to global feature aggregation. However, there is still a gap \wrt softmax attention, especially in the wide-baseline setting.
This suggests that future work should focus on reconciling the fixed expressivity of the MLP scene representation with the high accuracy of quadratic attention.

\PAR{Acknowledgments.}
We thank Alessandro Bursio for providing valuable feedback on the draft of this paper. We also thank Tobias Fischer and Alessandro Bursio for helping with the training and evaluation data setup used for this work.

%% file: sec/6_suppl.tex
\clearpage
\setcounter{page}{1}
\setcounter{section}{0}
\renewcommand{\thesection}{\Alph{section}}
\renewcommand{\theHsection}{supp.\thesection}
\maketitlesupplementary

\noindent In this appendix, we provide:
\begin{itemize}
\item \textbf{A detailed description of the implementation and training process} of \methodname, including dataset usage, image collection sampling (based on co-visibility), training hyperparameters, and the specific parameters that are optimized during training (\cref{sec:implementation_details}).
\item \textbf{Enhancements to the VGGT baseline} that enables processing of larger image collections as well as \textit{increased accuracy} making it a stronger baseline. (\cref{sup:sec:vggt_adjustment})
\item \textbf{Further ablation studies} on key components of our method, including the effect of the number of optimizer steps used for the Test-Time Training (TTT) objective and an investigation into different filter configurations for the ShortConv2D layer, showing optimal settings (\cref{sec:additional_results}).
\item \textbf{Additional qualitative results and visualizations} for comparison with baselines (VGGT, TTT3R), qualitative examples of visual localization, and a discussion of the method's performance on scenes with larger spatial extent (\cref{sec:additional_qualitative_results}).
\end{itemize}

\section{Implementation Details}
\label{sec:implementation_details}

\PAR{Training.} 
We list the datasets used for training in \cref{sec:sup:tbl:training_datasets}. To obtain an image collection during training, we follow a greedy sampling approach: The algorithm starts by randomly sampling the first image, then uniformly samples from the set of images with co-visibility greater than $0.3$ with any of the images currently in the collection. This step repeats until the desired collection size is reached. We pre-compute the required co-visibility matrix via a depth consistency check~\cite{sun_loftr_2021}.

Following VGGT, we use an adaptive batch size with image collections of 2-24 images while keeping the total number of images per GPU at approximately 48. The image aspect ratio is sampled uniformly from the interval $[0.5, 2.0]$, and images are then resized such that their longer side is $518$. During training, we apply color jitter augmentation to each image independently, making the network more robust to brightness and contrast changes.

We train \methodname using AdamW~\cite{loshchilov_decoupled_2019} with a learning rate of $10^{-4}$, weight decay of $0.05$, and $\beta_1=0.9,\beta_2=0.95$. The learning rate increases by a factor of $10$ during the first $1,000$ training steps, then decays following a cosine schedule to a final learning rate of $10^{-6}$. For the inner optimization of the test-time training objective, we use Muon~\cite{jordan2024muon} with 5 Newton-Schulz iterations, a learning rate of $0.1$, and employ $1$ optimizer step during training. The TTT MLPs use input and output dimension $1024$, matching the hidden state size of VGGT, and projects to $4\times$ the input dimension in their hidden layers.
We train only the QKV projection matrices as well as the output projection in the global attention layers and the newly introduced parameters of the TTT module, while keeping all remaining parameters of the VGGT architecture (including encoder, per-image attention, and prediction heads) frozen.

Additionally, only the values projected from image patch tokens participate in the ShortConv2D operation. The camera and register tokens are passed through.

\begin{table}[t]
\label{sec:sup:tbl:training_datasets}
\centering
\adjustbox{max width=\linewidth}{%
\begin{tabular}{ll}
\toprule
Type & Dataset \\
\midrule
\multirow{7}{*}{Indoor} & Aria Synthetic Environments~\cite{avetisyan_scenescript_2024} \\
& DynamicReplica~\cite{karaev_dynamicstereo_2023} \\
& Hypersim~\cite{roberts_hypersim_2021} \\
& Replica~\cite{straub_replica_2019} \\
& Cubify Anything~\cite{lazarow_cubify_2025} \\
& Scannet++~\cite{yeshwanth_scannet_2023} \\
& Scannet~\cite{dai_scannet_2017} \\
& Taskonomy~\cite{zamir_taskonomy_2018} \\
\midrule
\multirow{5}{*}{Outdoor} 
& Mapillary Metropolis~\cite{research_mapillary_nodate} \\
& MatrixCity~\cite{li_matrixcity_2023} \\
& Megadepth~\cite{li_megadepth_2018} \\
& Mid-Air~\cite{fonder_mid-air_2019} \\
& Mapillary Planet-scale Depth Dataset~\cite{antequera_mapillary_2020} \\
& ParallelDomain4D~\cite{van_hoorick_generative_2025,parallel_domain} \\
& vKITTI2~\cite{gaidon_virtual_2016,cabon_virtual_2020} \\
\midrule
\multirow{3}{*}{Object centric} & CO3Dv2~\cite{reizenstein_common_2021} \\
& Kubric~\cite{greff_kubric_2022} \\
& Wild-RGBD~\cite{xia_rgbd_2024} \\
\midrule
\multirow{4}{*}{Mixed} & BlendedMVG~\cite{yao_blendedmvs_2020} \\
& DL3DV-10K~\cite{ling_dl3dv-10k_2024} \\
& Spring~\cite{mehl_spring_2023} \\
& TartanAirV2~\cite{wang_tartanair_2020} \\
& UnrealStereo4k~\cite{tosi_smd-nets_2021} \\
\bottomrule
\end{tabular}
}
\caption{Datasets used for training.}
\end{table}

\PAR{Inference details.}
The VGGT architecture, which we initialize with, has multiple decoders that predict redundant geometric quantities. To obtain pointmaps one can either use the outputs of the global pointmap prediction head directly or use the camera and depth predictions together to unproject to pointmaps. While VGGT finds the latter to be more precise, we use the global prediction head to obtain pointmaps due to the imprecise camera pose predictions mentioned in \cref{sec:experiments:subsec:standard_benchmarks}, which would otherwise degrade the pointmaps obtained by unprojecting depth.
In \cref{sec:experiments:subsec:visloc}, we retain the camera tokens of all mapping images as input to the camera head in the visual localization setting since VGGT's camera head requires the camera tokens of all images. The camera token of the query image then participates in the softmax attention operation in the camera head before it is decoded to camera parameters.
For all benchmarking, we use NVIDIA A100-80GB GPUs.

\PAR{Further evaluation details.}
For visual localization results in \cref{sec:experiments:subsec:visloc}, we sub-sample mapping images at a stride of 200 for 7Scenes and 20 for Wayspots.
For pointmap evaluation \cref{sec:experiments:subsec:large_scale}, the usage of the iterative closest point (ICP) algorithm for alignment of prediction and ground truth point clouds makes evaluation very slow when evaluating predictions on large image sets. We instead select a set of equally spaced keyframes, that capture the scene geometry, to compute pointmap metrics while we treat all other frames as supporting views. For our evalution we use 10 keyframes. For TTT3R, we provide the images in sequential order with the keyframes last such that the model has seen all images of the scene before making predictions.

\section{VGGT adjustments}
\label{sup:sec:vggt_adjustment}
To enable a fair comparison with VGGT in the setting with a large number of images in \cref{sec:experiments:subsec:large_scale}, we perform several changes in the VGGT codebase that enhance its performance. 

\PAR{Memory-optimizations and distributed inference.}
First, we follow \citet{shen_fastvggt_2025} and discard unused activations in VGGT's alternating attention module, which allows processing up to $1k$ images on a single 80GB GPU. Next, we enable context parallel inference using Ulysses~\cite{jacobs_deepspeed_2023}, implemented in TransformerEngine~\footnote{\url{https://docs.nvidia.com/deeplearning/transformer-engine/user-guide/api/pytorch.html\#transformer_engine.pytorch.DotProductAttention}}, in the global attention layers. We note that the underlying attention implementation still uses FlashAttention2~\cite{dao_flashattention_2022}. While this allows VGGT to run for $2k$ images, as we show in \cref{sec:experiments:subsec:large_scale}, this requires runtimes up to $47$ minutes on 2 GPUs.

\PAR{Enhanced long-sequence generalization.}
For fair comparison on large image collections, we further adjust the scale parameter of the softmax in the global attention layers similar to the approach of \citet{jin_training-free_2023}, ensuring the entropy of the attention matrix stays constant. Let
\begin{equation}
    a_{i,j} = \frac{\text{exp}(\lambda k_i^T q_j)}{\sum_k \exp(\lambda k_k^T q_k)} 
\end{equation}
be the attention scores as used in softmax attention where $\lambda = \nicefrac{1}{\sqrt{d}}$~\cite{vaswani_attention_2017}. We instead set
\begin{equation}
    \lambda^\prime = \lambda \max(1.0, \log_{N_T} N),
\end{equation}
where $N_T$ and $N$ are the maximum number of tokens seen during training and of the current sequence, respectively. This ensures that the scaling is the same for sequence lengths seen during training, while for larger sequence lengths the attention matrix is sharpened. Since VGGT trains using a maximum of 24 images with $518\times518$ resolution and uses patch size $14$, we set $N_T = 24 * (518 / 14)^2 = 32,856$. We show improved performance using this entropy-scaling in \cref{sec:sup:tab:vggt_entropy} for large image collections making the VGGT baseline significantly stronger.

\begin{table}[t]
    \centering
    \adjustbox{max width=\linewidth}{%
    \input{tables/sup_vggt_entropy_scaling}
    }
    \caption{Attention entropy-scaling makes VGGT a stronger baseline on large image collections.}
    \label{sec:sup:tab:vggt_entropy} 
\end{table}

\section{Additional Results}
\label{sec:additional_results}

\begin{figure}[t]
    \centering
    \includegraphics[width=\linewidth]{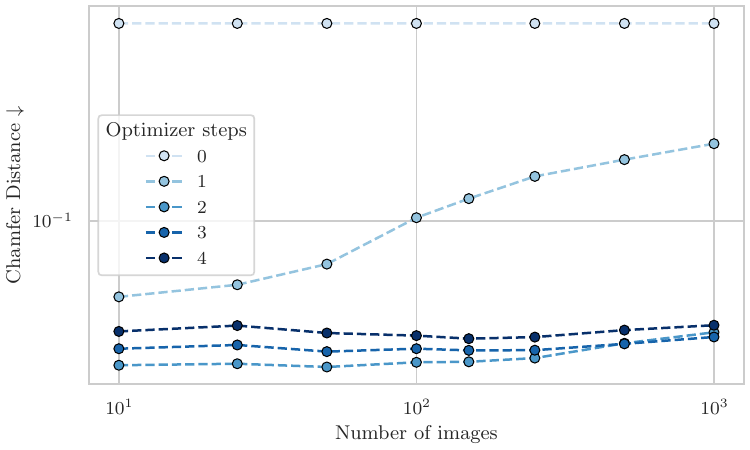}
    \caption{Pointmap error with increasing number of images when varying the optimizer steps on the TTT objective.}
    \label{sec:experiments:fig:ablation_num_steps}
\end{figure}

\PAR{Number of optimizer steps.}
We provide an additional evaluation varying the number of steps used to optimize the TTT objective~\cref{sec:method:subsec:scene_rep:eq:ttt_opt_1} at inference time for varying image collection sizes. We report results on the NRGBD dataset~\cite{azinovic_neural_2022} in \cref{sec:experiments:fig:ablation_num_steps}. As expected, we find that without TTT optimization the reconstruction error is high as no global information is propagated across tokens. A single optimizer step is sufficient for image collection sizes seen during training; however, the reconstruction error degrades as the number of images extends beyond that. Two optimizer steps achieve the best performance across a wide range of image collection sizes, and further increasing the number of steps to 3 or 4 leads to comparable or slightly worse performance.

\PAR{ShortConv2D.}
In the main paper, we find improved performance when using a $3\times3$ ShortConv2D on values $v_i$ of the attention operation before optimizing the MLP using TTT. 
Here, we provide further experiments using different configurations of our ShortConv2D. In addition to a $3\times3$ filter on the values $v_i$ ($V\text{-}3$) used in the main paper, we consider a $5\times5$ filter ($V\text{-}5$) and a variant where we apply ShortConv2D to keys $k_i$ and values $v_i$ jointly ($KV\text{-}3$).

We report results in \cref{sec:sup:tab:shortconv2d}. We observe that increasing the filter size from $3$ to $5$ does not further increase performance, showing that a filter size of $3$ is sufficient to obtain a strong self-supervised objective for TTT. Applying ShortConv2D to both the keys and values results in decreased performance. We explain this by the fact that applying the same spatial mixing does not break the dependency between keys and values, as explained in \cref{sec:experiments:subsec:large_scale}.

\begin{table}[t]
    \centering
    \adjustbox{max width=\linewidth}{%
    \input{tables/sup_shortconv2d}
    }
    \caption{\textbf{Results for different filter configuration in ShortConv2D.}}
    \label{sec:sup:tab:shortconv2d} 
\end{table}

\section{Additional Qualitative Results}
\label{sec:additional_qualitative_results}

\begin{figure*}[p]
    \centering
    \begin{subfigure}{\textwidth}
        \centering
        \includegraphics[width=0.8\textwidth]{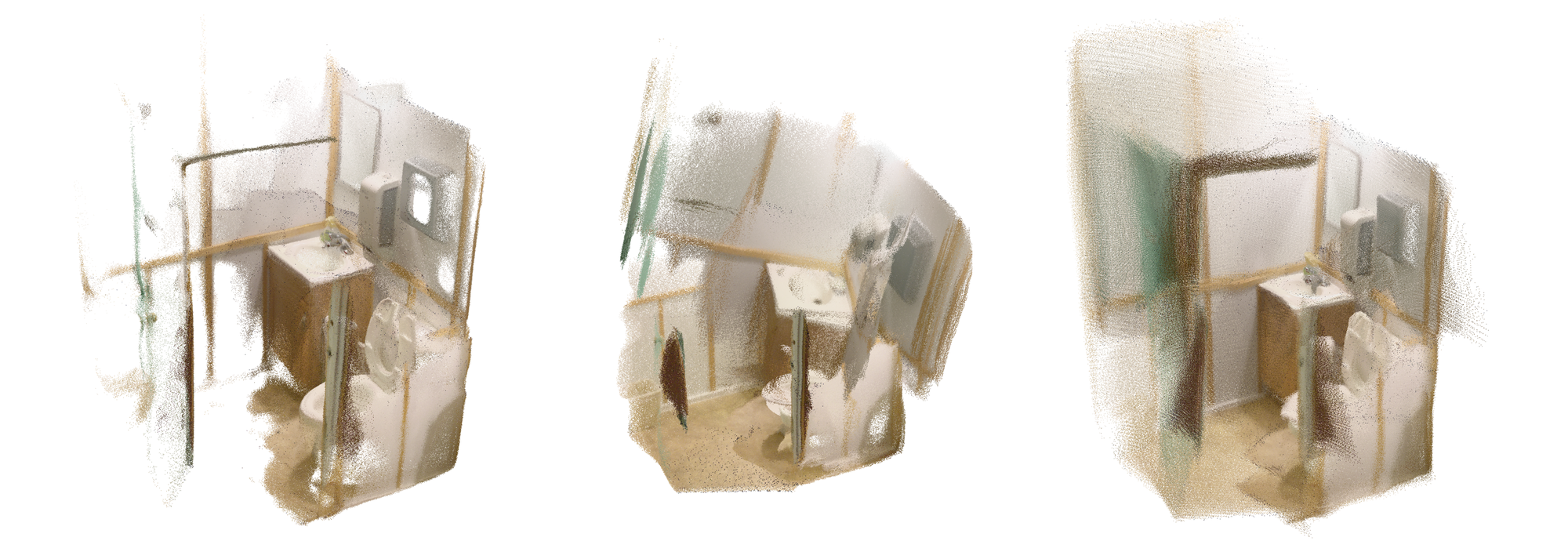}
    \end{subfigure} \\
    \begin{subfigure}{\textwidth}
        \centering
        \includegraphics[width=0.8\textwidth]{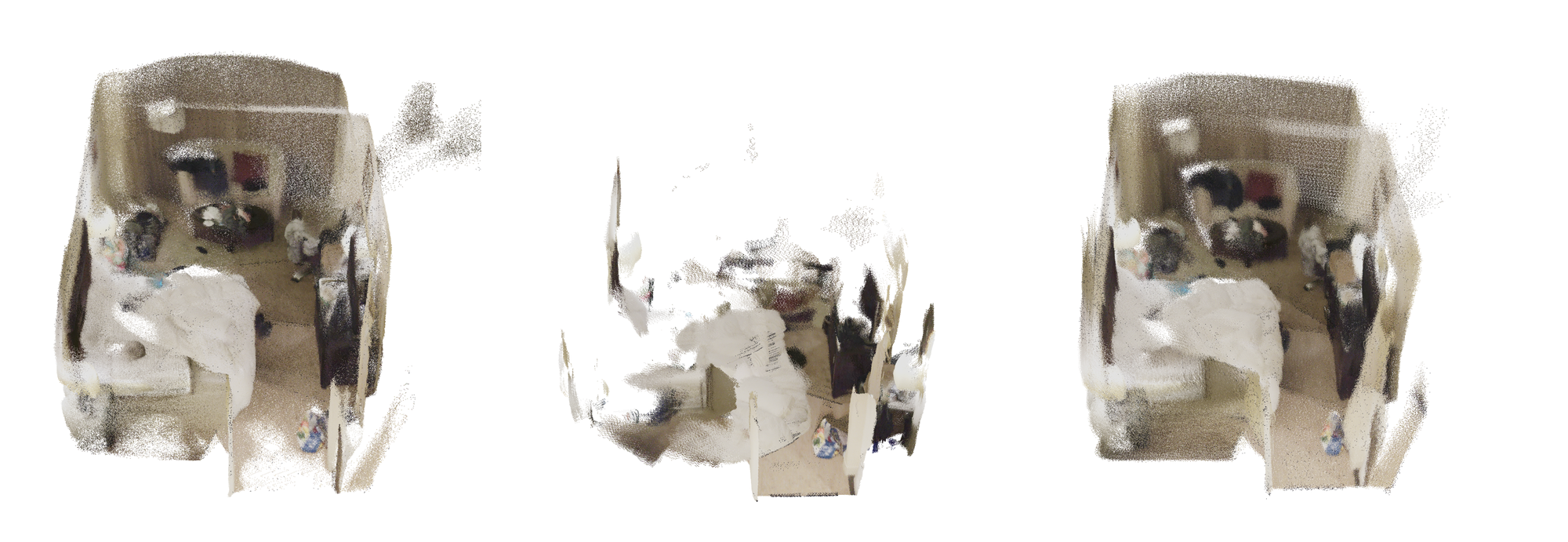}
    \end{subfigure} \\
    \begin{subfigure}{\textwidth}
        \centering
        \includegraphics[width=0.8\textwidth]{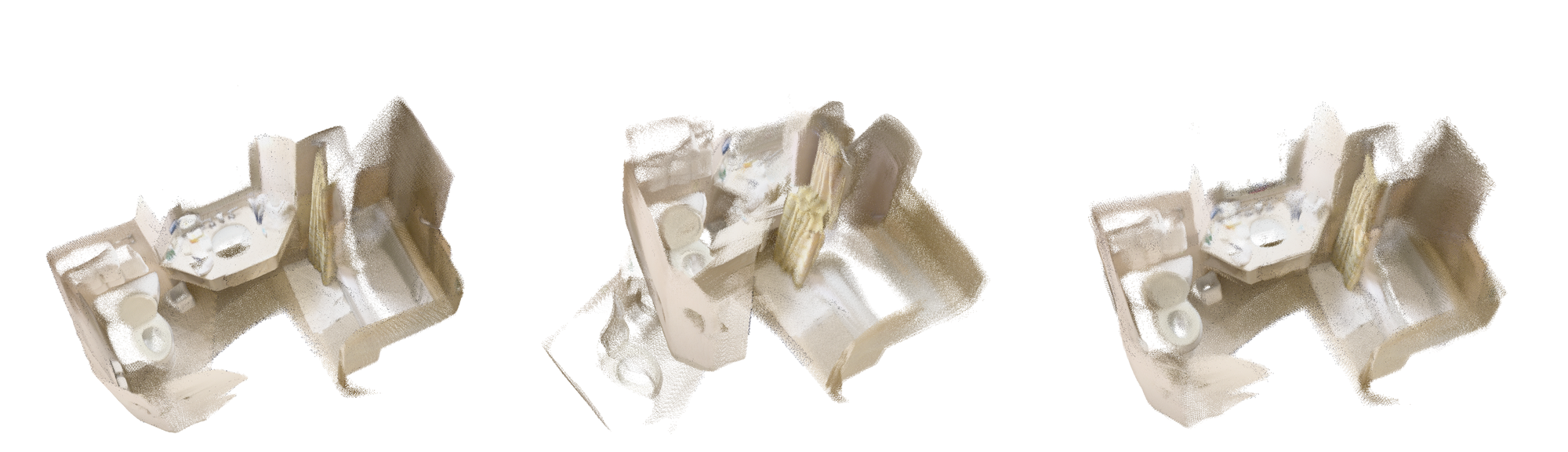}
    \end{subfigure} \\
    \begin{subfigure}{\textwidth}
        \centering
        \includegraphics[width=0.8\textwidth]{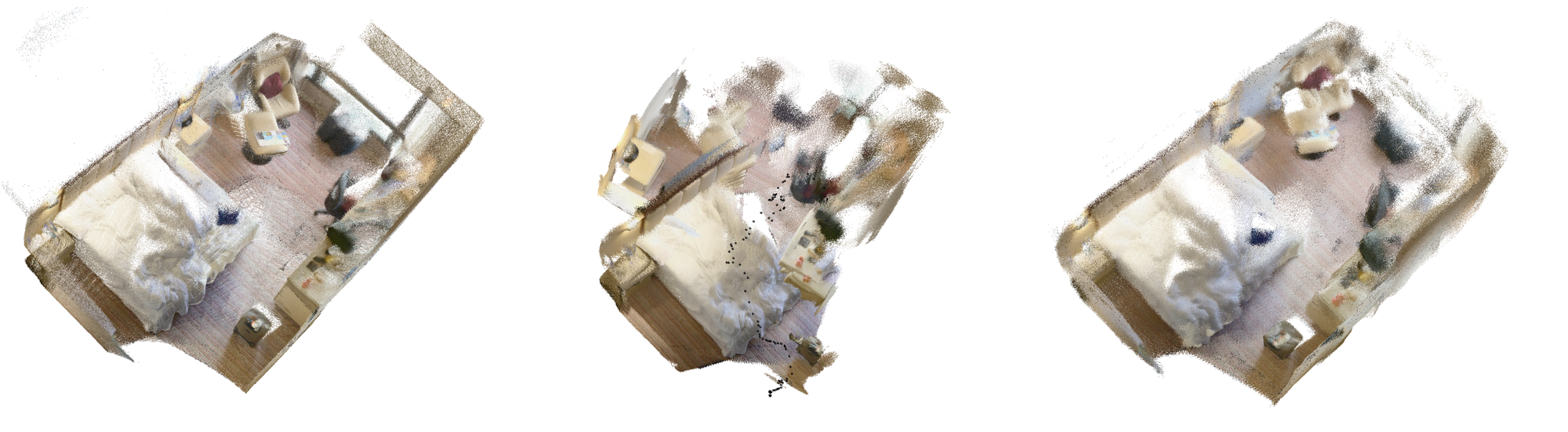}
    \end{subfigure} \\
    \begin{subfigure}{\textwidth}
        \centering
        \includegraphics[width=0.8\textwidth]{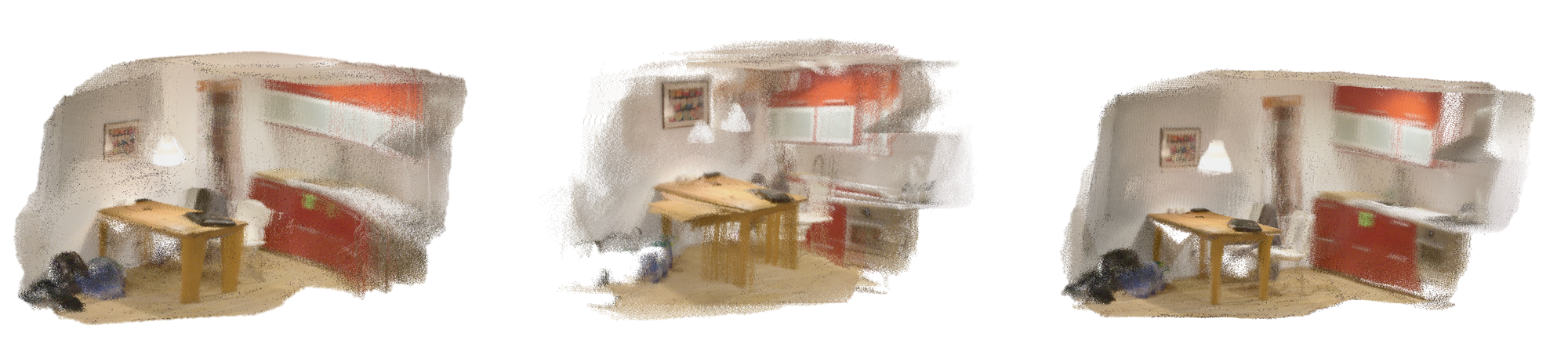}
    \end{subfigure}
    \caption{\textbf{Qualitative comparison.} From left to right: VGGT, TTT3R, \methodname (Ours)}
    \label{sec:supl:fig:qualitative_scannet}
\end{figure*}

\PAR{Qualitative comparison.}
We report additional qualitative comparisons between VGGT, TTT3R, and \methodname in \cref{sec:supl:fig:qualitative_scannet}. TTT3R and \methodname process these $1k$ image collections within $1$ minute; however, our method produces 3D consistent reconstructions while TTT3R degrades significantly. VGGT achieves slightly sharper details but takes more than $11$ minutes due to the quadratic scaling of softmax attention.

\begin{figure*}[p]
    \centering
    \begin{subfigure}{\textwidth}
        \centering
        \includegraphics[width=0.7\textwidth]{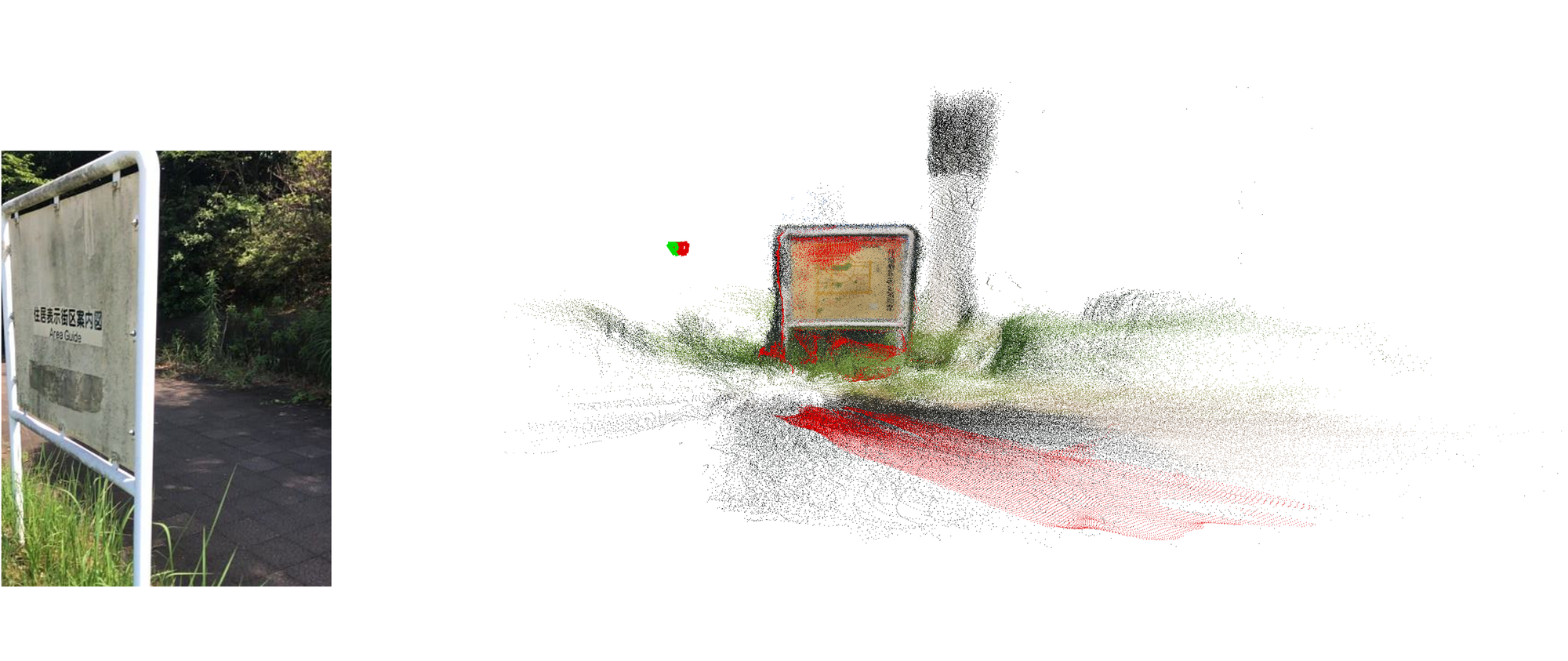}
    \end{subfigure} \\
    \begin{subfigure}{\textwidth}
        \centering
        \includegraphics[width=0.7\textwidth]{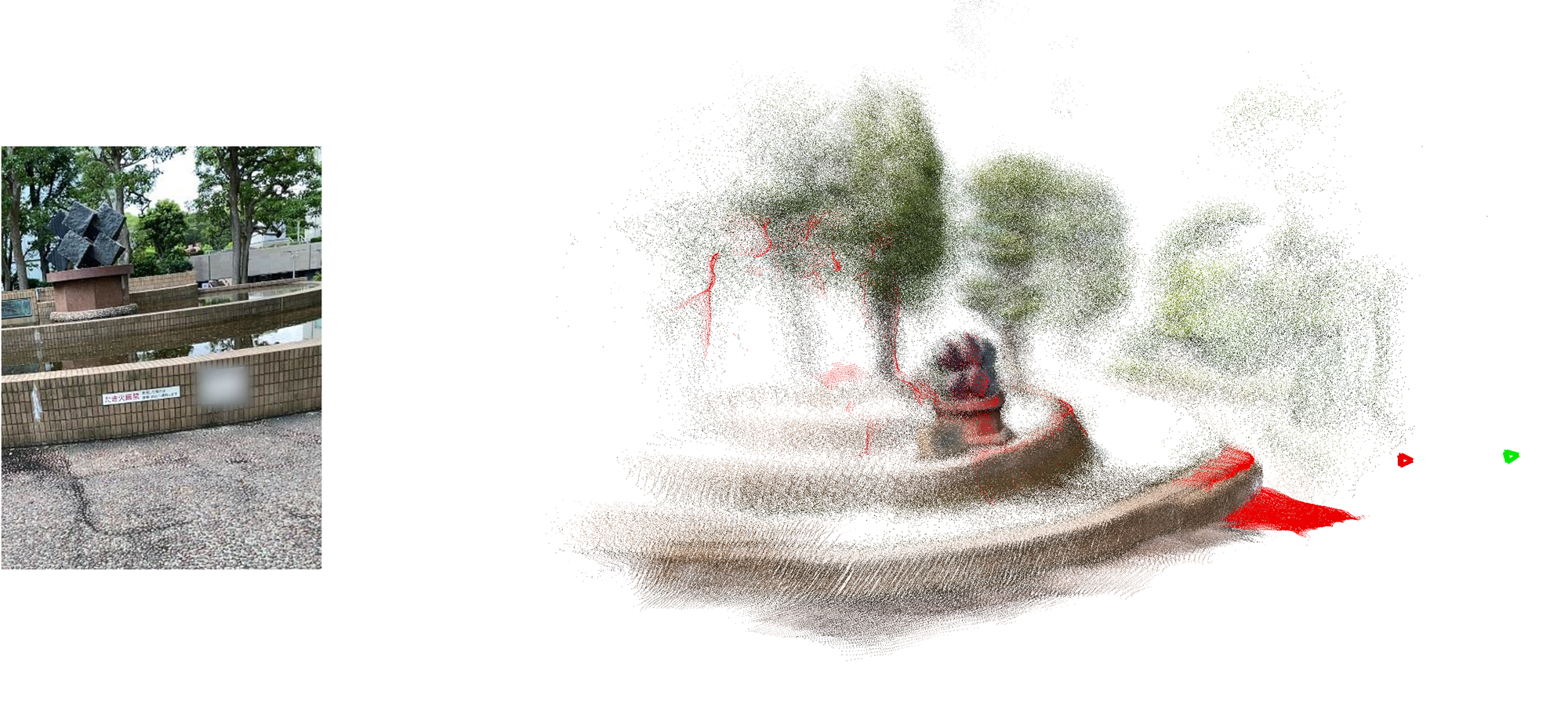}
    \end{subfigure} \\
    \begin{subfigure}{\textwidth}
        \centering
        \includegraphics[width=0.7\textwidth]{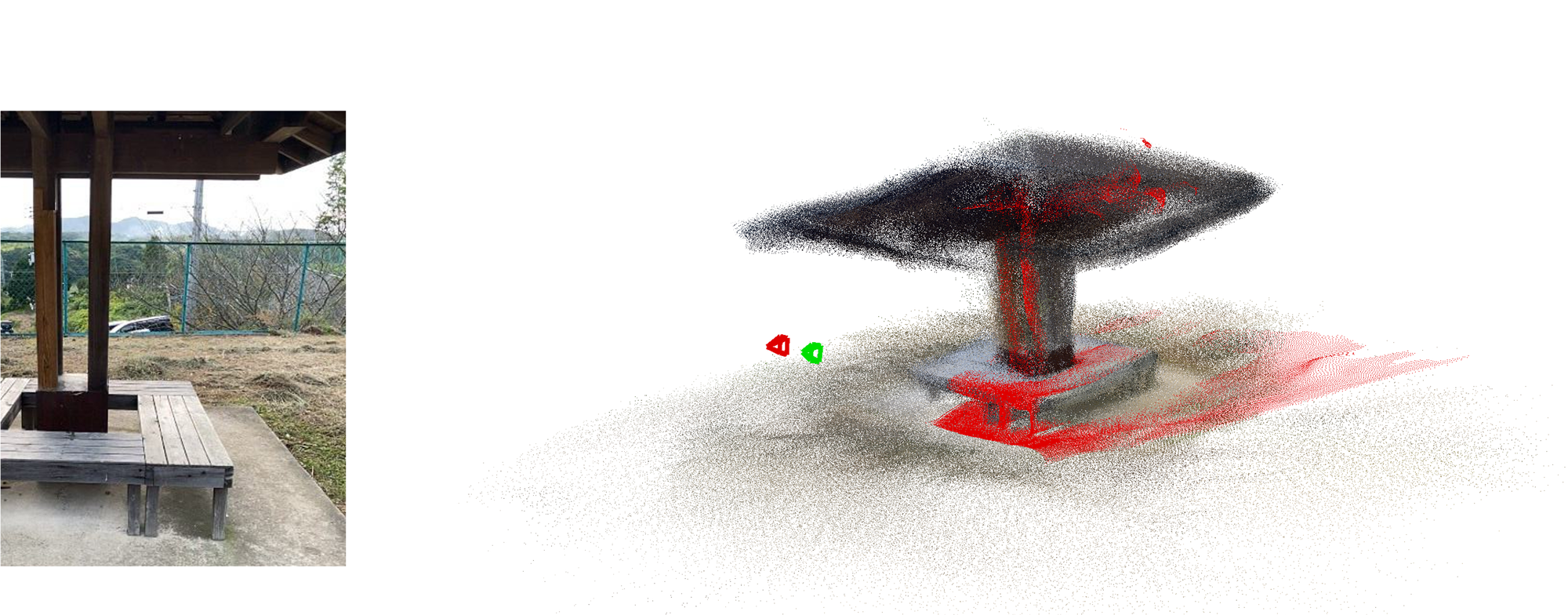}
    \end{subfigure} \\
    \begin{subfigure}{\textwidth}
        \centering
        \includegraphics[width=0.7\textwidth]{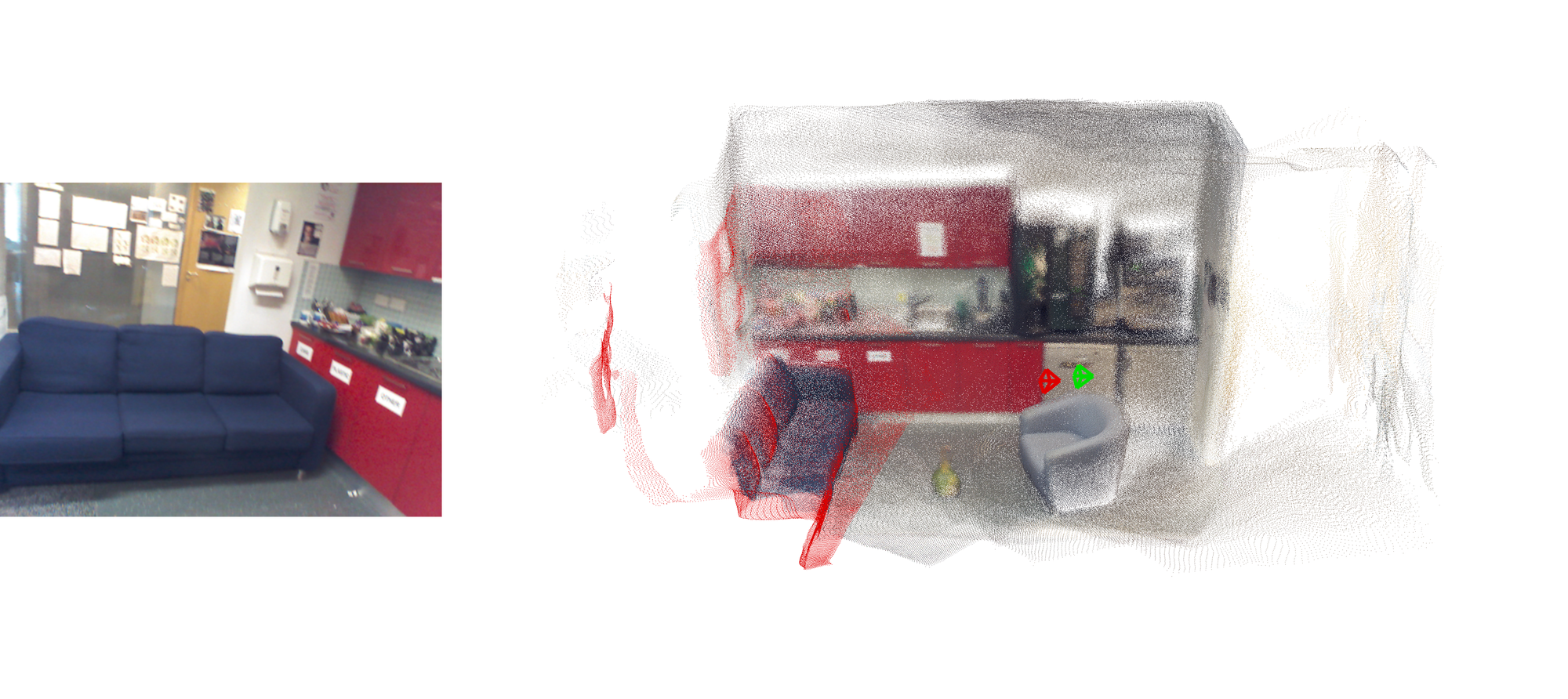}
    \end{subfigure} \\
    \caption{\textbf{Visual localization examples in Wayspots and 7scenes.} Ground truth camera for query image (not used for reconstruction) shown on the left in green, predicted camera and geometry in red.}
    \label{sec:supl:fig:visloc_qualitative}
\end{figure*}

\begin{figure*}[t]
    \centering
    \includegraphics[width=0.7\textwidth]{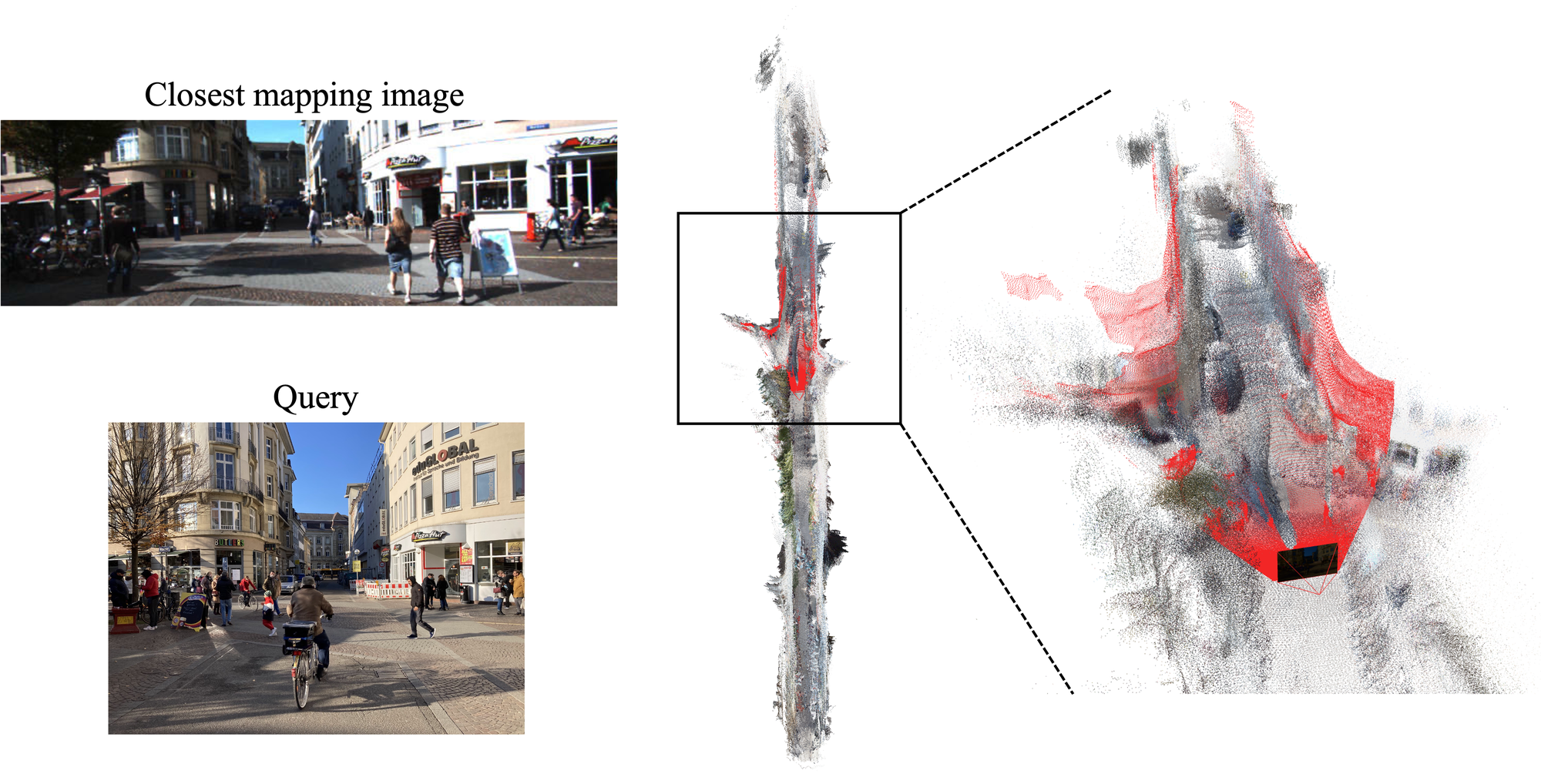}
    \caption{\textbf{In-the-wild visual localization.} We reconstruct a sequence of the KITTI dataset, then localize a tourist picture that was recorded 7 years later. Note the changes in appearance and composition of the scene.}
    \label{sec:supl:fig:visloc_in_the_wild}
\end{figure*}

\PAR{Visual localization examples.}
Complementary to the visual localization results in \cref{sec:experiments:subsec:visloc}, we show examples of localizing query images in the completed reconstruction by running the frozen MLPs in \cref{sec:supl:fig:visloc_qualitative}.
In \cref{sec:supl:fig:visloc_in_the_wild}, we show an in-the-wild example where we localize a tourist picture taken from a phone camera, together with its geometry, within a recording of an autonomous vehicle from the KITTI dataset that is 7 years older. Despite the temporal gap and changes in the street, our method successfully localizes the query image. We observe that the tourist photo captures upper parts of buildings not visible from the car-mounted camera, demonstrating the robustness of our approach to viewpoint variations.

\begin{figure*}[p]
    \centering
    \begin{subfigure}{\textwidth}
        \centering
        \includegraphics[width=0.8\textwidth]{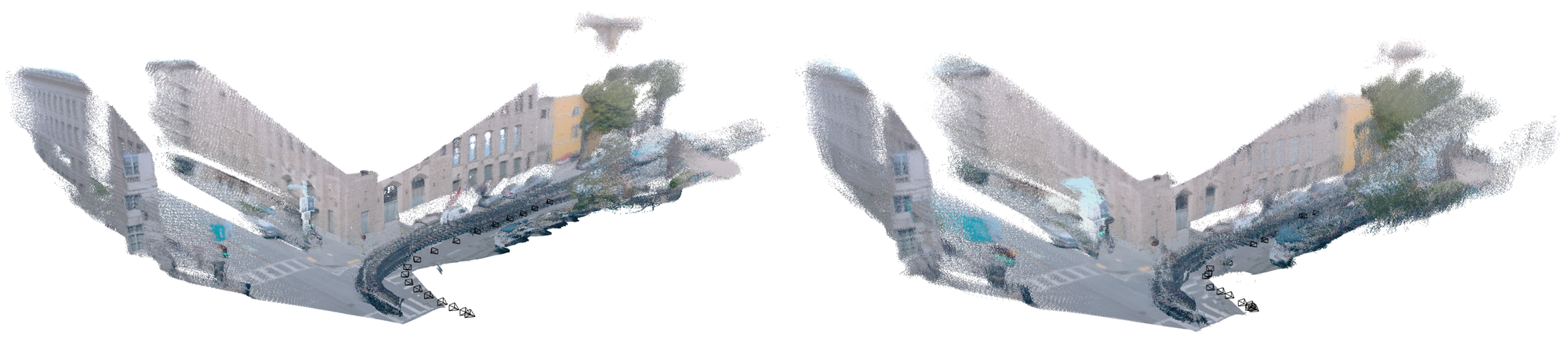}
    \end{subfigure} \\
    \begin{subfigure}{\textwidth}
        \centering
        \includegraphics[width=0.8\textwidth]{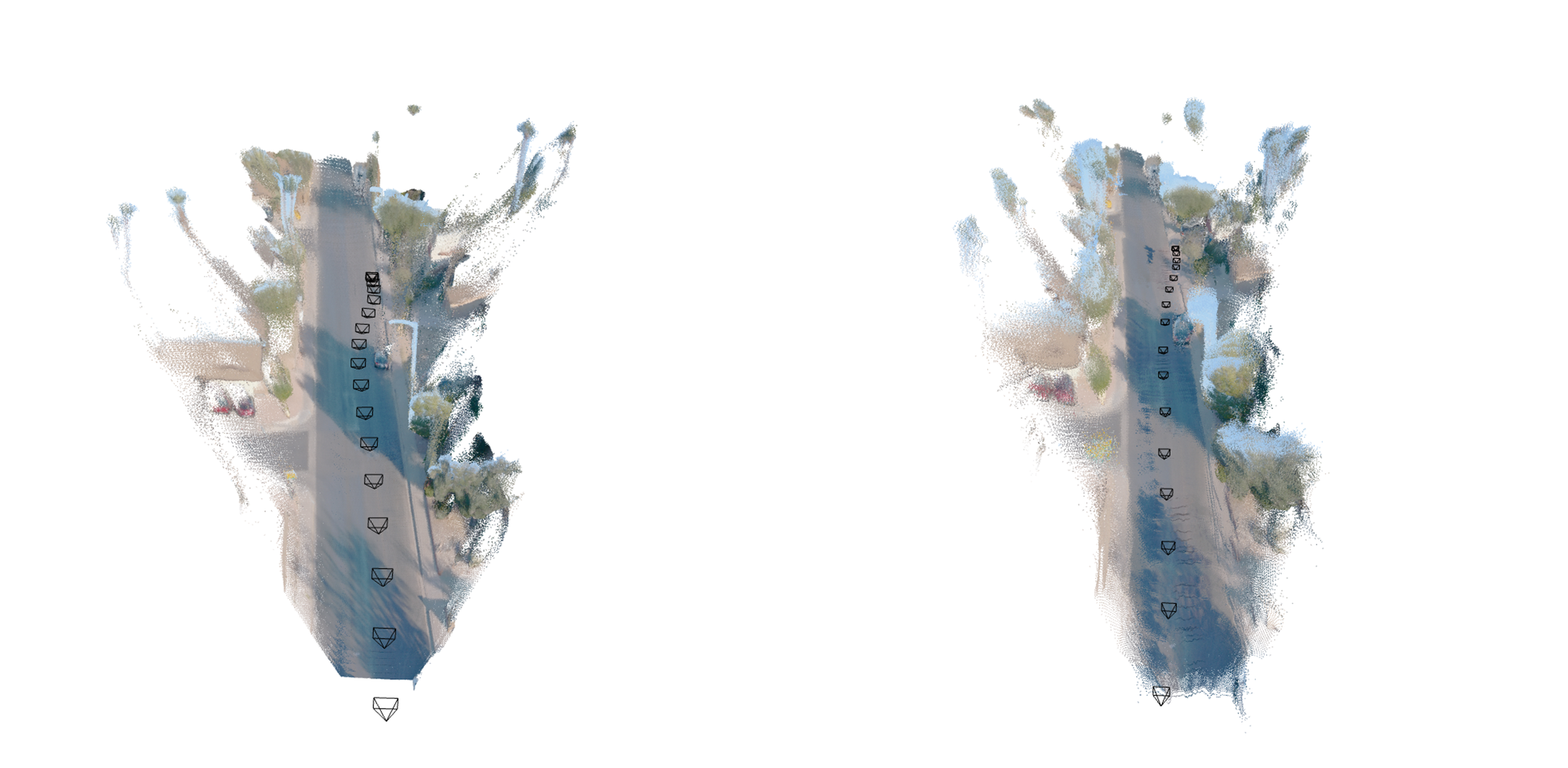}
        \caption{Similar reconstruction as VGGT.}
    \label{sec:supl:subfig:waymo_success}
    \end{subfigure} \\
    \begin{subfigure}{\textwidth}
        \centering
        \includegraphics[width=0.8\textwidth]{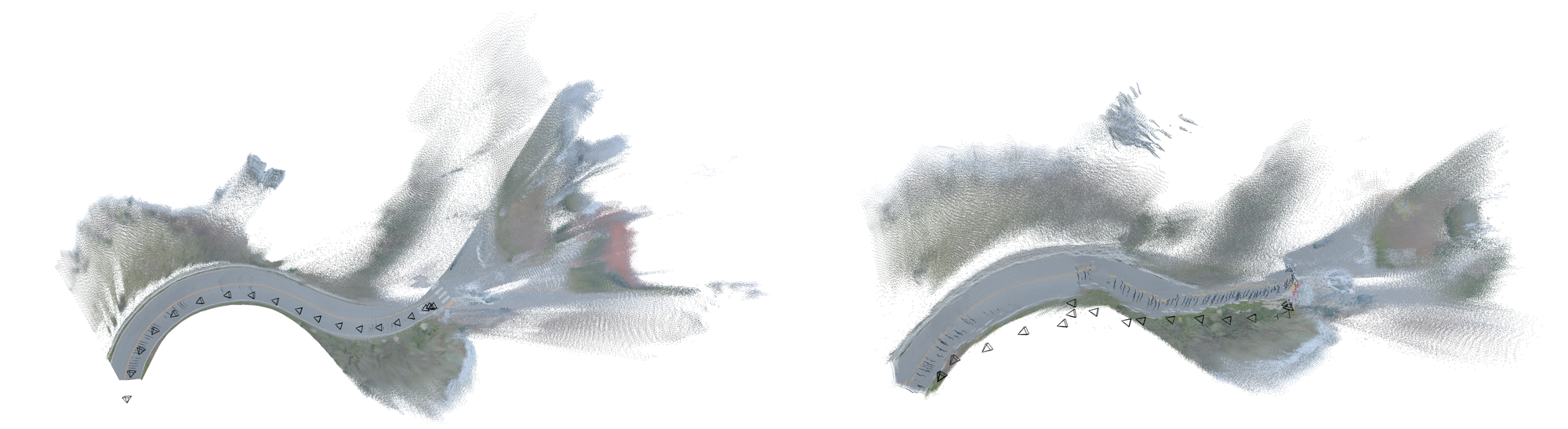}
    \end{subfigure} \\
    \begin{subfigure}{\textwidth}
        \centering
        \includegraphics[width=0.8\textwidth]{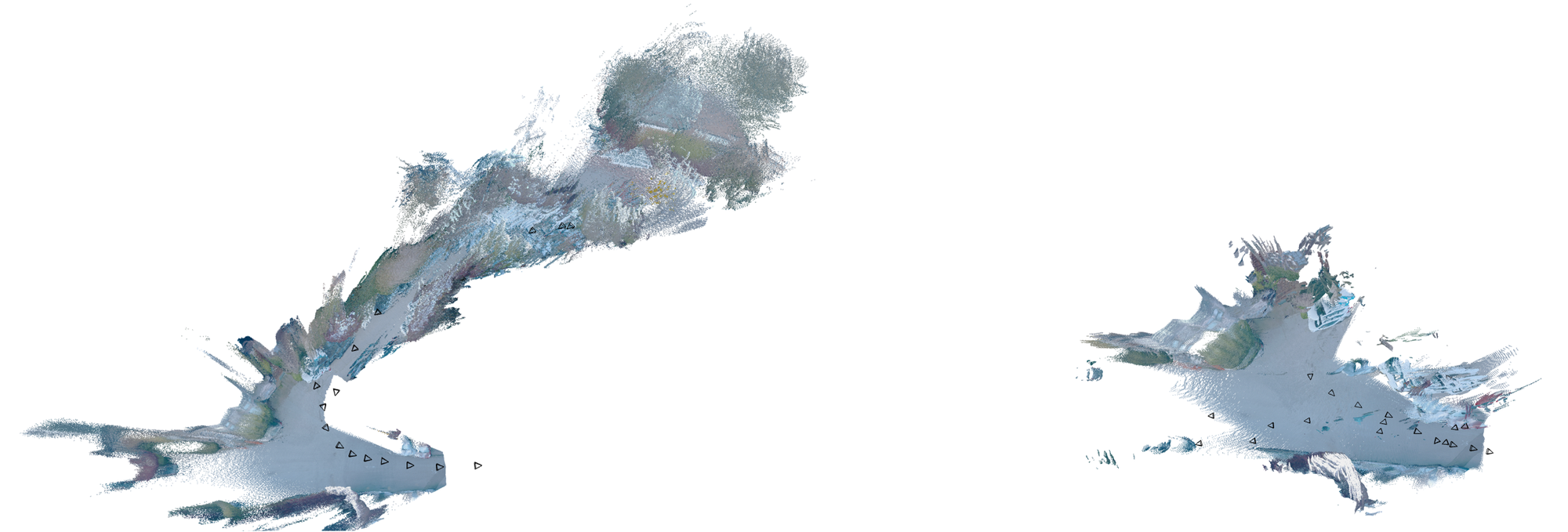}
        \caption{Failure cases.}
    \label{sec:supl:subfig:waymo_failures}
    \end{subfigure} \\
    \caption{Waymo sequence reconstructions comparison with VGGT.}
    \label{sec:supl:fig:waymo}
\end{figure*}

\PAR{Scenes with larger spatial extent.} 
We visualize reconstructions of Waymo sequences that have larger spatial extent in \cref{sec:supl:fig:waymo}. While \methodname can often achieve similar results to VGGT (\cref{sec:supl:subfig:waymo_success}), in some cases with more complex scene layouts, the reconstruction quality is degraded (\cref{sec:supl:subfig:waymo_failures}). We note this as a limitation that linear-time attention mechanisms cannot yet match softmax attention in all cases; however, this also provides an interesting avenue to explore for future work by, \eg, adapting the amount of computation depending on scene complexity and designing more expressive linear attention mechanisms that match the accuracy of softmax attention.

%% file: tables/sup_vggt_entropy_scaling.tex
\begin{tabular}{lrrrrrrrr}
\toprule
\#images & \multicolumn{2}{c}{250} & \multicolumn{2}{c}{500} & \multicolumn{2}{c}{750} & \multicolumn{2}{c}{1000} \\
\cmidrule(lr){2-3} \cmidrule(lr){4-5} \cmidrule(lr){6-7} \cmidrule(lr){8-9} 
 & CD $\downarrow$ & NC $\uparrow$ & CD $\downarrow$ & NC $\uparrow$ & CD $\downarrow$ & NC $\uparrow$ & CD $\downarrow$ & NC $\uparrow$ \\
\midrule
VGGT & 0.018 & \textbf{0.894} & 0.025 & 0.876 & 0.040 & 0.864 & 0.041 & 0.855 \\
VGGT + Entropy-scaling & \textbf{0.016} & \textbf{0.894} & \textbf{0.017} & \textbf{0.889} & \textbf{0.030} & \textbf{0.871} & \textbf{0.029} & \textbf{0.872} \\
\bottomrule
\end{tabular}

%% file: tables/sup_shortconv2d.tex
\begin{tabular}{lrrr}
\toprule
 & CD $\downarrow$ & NC $\uparrow$ & mAA(30) $\uparrow$ \\
\midrule
No ShortConv2D & 0.074 & 0.833 & 72.16 \\
$V\text{-}3$ & \textbf{0.066} & \textbf{0.838} & \textbf{74.14} \\
$V\text{-}5$ & 0.069 & 0.833 & 72.52 \\
$K\text{-}3$ & 0.068 & 0.834 & 72.89 \\
$KV\text{-}3$ & 0.081 & 0.820 & 69.44 \\
\bottomrule
\end{tabular}